%% file: main.tex
\documentclass[final,5p,times,authoryear]{elsarticle}

\makeatletter
\renewcommand\subsection{\@startsection{subsection}{2}{\z@}%
           {12\p@ \@plus 6\p@ \@minus 3\p@}%
           {3\p@ \@plus 6\p@ \@minus 3\p@}%
           {\normalfont\normalsize\bfseries\boldmath}}
\renewcommand\subsubsection{\@startsection{subsubsection}{3}{\z@}%
           {12\p@ \@plus 6\p@ \@minus 3\p@}%
           {\p@}%
           {\normalfont\normalsize\bfseries\boldmath}}
\makeatother

\usepackage{amsmath}
\usepackage{amssymb}
\usepackage{graphicx}
\usepackage{booktabs}
\usepackage{longtable}
\usepackage[hidelinks]{hyperref}

\makeatletter
\def\ps@pprintTitle{\let\@oddhead\@empty\let\@evenhead\@empty
  \let\@oddfoot\@empty\let\@evenfoot\@empty}
\makeatother

\begin{document}

\begin{frontmatter}

\title{From greenhouse climate to individual leaves: an organ-resolved model of lettuce growth}

\author[bse]{Md Hasibur Rahman}
\ead{mzr0134@auburn.edu}

\author[bse]{Faraz Ahmed}
\ead{fza0070@auburn.edu}

\author[hort]{Hafiz Muhammad Bilal}

\author[hort]{Daniel Wells}

\author[synap]{Dylan Tobin}
\ead{dylant@synapgarden.com}

\author[bse,hort]{Tanzeel U. Rehman\corref{cor1}}
\ead{tur0001@auburn.edu}
\cortext[cor1]{Corresponding author.}

\address[bse]{Department of Biosystems Engineering, Auburn University, Auburn,
AL 36849, United States}
\address[hort]{Department of Horticulture, Auburn University, Auburn, AL 36849,
United States}
\address[synap]{SynapGarden, Detroit, MI 48226, United States}

\begin{abstract}
Greenhouse climate management aims to improve crop production while limiting
energy use. This requires knowing how a crop may respond before greenhouse
conditions are changed. A crop digital twin can support this decision, but it
must represent how plant physiology and structure develop together. In this
study, a unified framework was developed to simulate lettuce growth using
physiological information from individual leaves. Each leaf received the
environmental conditions at its position in the canopy and contributed carbon
through photosynthesis. Part of this carbon was used for maintenance, while
the remainder supported growth and was distributed among leaves based on their
age, size and local environment. The predicted leaf mass, area
and age generated an evolving three-dimensional plant in NVIDIA Isaac Sim. Ray
tracing calculated the radiation intercepted by each leaf and returned this
information to photosynthesis, allowing plant structure and growth to influence
each other over time. In the comparison with greenhouse measurements, the
relative root mean square error was 9.5\% for total dry weight and 9.2\%,
12.7\% and 13.1\% for leaf number, canopy diameter and largest-leaf area,
respectively. A 30\% decrease in incident radiation reduced final dry weight by 10.4\%,
while the same increase raised it by 6.9\%, and adding 200~ppm carbon dioxide
raised it by 46.1\%. Within a simulated 40-plant block, interior plants
accumulated 8.6\% less dry weight than border plants with identical initial
states, and the leaf-specific tipburn index rose in the enclosed leaves over
the same period in which tipburn appeared on the greenhouse plants. These
results show that resolving individual leaves
can explain how local exposure changes plant growth within the greenhouse. The
unified framework provides the forward plant model needed for a future
bidirectional digital twin, where observations of the physical plant can update
predictions and support greenhouse climate decisions.
\end{abstract}


\end{frontmatter}

\section{Introduction}
\label{sec:introduction}
\input{sections/new_intro}

\section{Materials and methods}
\label{sec:methods}
\input{sections/2_1_overview}

\subsection{Development of a leaf-level lettuce growth model}
\label{sec:leaf_level_growth_model}
\input{sections/2_2_1_plant_level}
\input{sections/2_3_leaf_dynamics}
\input{sections/2_2_2_organ_resolved}

\subsection{Development of the functional-structural plant model}
\label{sec:functional_structural_model}
\input{sections/2_4_structure}

\subsection{Simulation in the NVIDIA Isaac Sim greenhouse environment}
\label{sec:isaac_sim_environment}
\input{sections/2_5_1_radiation}
\input{sections/2_5_2_microclimate}
\input{sections/2_6_leaf_physiology}
\input{sections/2_7_tipburn}

\subsection{Numerical implementation}
\label{sec:numerical_implementation}
\input{sections/2_9_implementation}

\subsection{Experimental design, calibration and evaluation}
\label{sec:calibration_and_evaluation}
\input{sections/plant_materials}
\input{sections/2_8_calibration}

\section{Results and discussion}
\label{sec:results}
\input{sections/results_1_3}
\input{sections/results_4_6}

\section{Conclusion}
\label{sec:conclusion}

This study developed a unified framework that connected greenhouse climate,
leaf-level physiology and three-dimensional lettuce development within NVIDIA
Isaac Sim. Growth emerged from the light received, the carbon assimilated and
the dry matter allocated leaf by leaf, with the carbon balance accounting for
maintenance in between, while the evolving canopy altered radiation
interception and subsequent growth. The model followed the measured progression
of biomass and structural development, with relative errors of 9.2 to 13.1\%
for dry weight, leaf number, canopy diameter and largest-leaf area. The
dry-weight error was of the same order as the 10.5 to 24.9\% reported for the
plant-level model the framework was built on, and leaf number and blade area
were comparable to a model dedicated to leaf development, in each case obtained
under different trials and sample sizes. Environmental and spatial simulations showed how
changes in climate and neighboring plants affected carbon assimilation, leaf
exposure and biomass accumulation, with the largest response to carbon dioxide
enrichment, and the leaf-specific tipburn index rose in the enclosed leaves over
the same period in which tipburn appeared on the greenhouse plants. Together,
these results demonstrated the ability of the framework to connect plant growth
with the local conditions experienced by individual leaves.

The framework was configured for one lettuce cultivar, and its parameters
remained fixed during the season rather than being updated from observations
of the corresponding physical plant. The tipburn index was compared with the
greenhouse plants on the timing of injury and on the leaves at risk, and a
plant-by-plant comparison is the next step once trial and simulated plants are
matched in size.

Future work will extend the framework to multiple cultivars and introduce
learnable physiological and structural parameters. An observation-driven
backward pass will update these parameters and the digital plant state from
repeated measurements of the physical plant, allowing subsequent predictions to
reflect observed development. Further development will establish a physics-based,
simulation-ready greenhouse for evaluating interactions between robots and
growing plants. These extensions will allow the unified framework to support
climate strategies that meet physical crop requirements while reducing greenhouse
energy demand, and to test robotic harvesting, grasp planning, plant detection,
leaf segmentation and three-dimensional phenotyping within the simulated
environment.

\section*{Data availability}

The data that support the findings of this study are available from the
corresponding author upon reasonable request.

\section*{Declaration of generative AI and AI-assisted technologies in the writing process}

During the preparation of this work the author(s) used Grammarly and ChatGPT to
correct grammar, enhance readability, and ensure a smooth flow of information.
After using this tool/service, the author(s) reviewed and edited the content as
needed and take(s) full responsibility for the content of the publication.

\bibliographystyle{elsarticle-harv}
\bibliography{references}

\clearpage
\appendix
\setcounter{figure}{0}
\setcounter{table}{0}
\providecommand{\theHsection}{}\renewcommand{\theHsection}{appx.\Alph{section}}
\providecommand{\theHsubsection}{}\renewcommand{\theHsubsection}{appx.\Alph{section}.\arabic{subsection}}
\providecommand{\theHfigure}{}\renewcommand{\theHfigure}{appx.\Alph{section}.\arabic{figure}}
\providecommand{\theHtable}{}\renewcommand{\theHtable}{appx.\Alph{section}.\arabic{table}}
\section{Supporting information}
\label{app:supporting}

\setlength{\emergencystretch}{4em}

\newcommand{\suppsec}[1]{\subsection{#1}}
\input{sections/supplement_results_4_6}

\end{document}

%% file: sections/new_intro.tex
Controlled environment agriculture (CEA) enables year-round production of leafy vegetables by regulating radiation, temperature, relative humidity, carbon dioxide concentration, and air movement. This control can substantially improve productivity. A meta-analysis of lettuce production reported an average yield of 3.68~kg~m$^{-2}$ in CEA systems compared with 1.88~kg~m$^{-2}$ in field production \citep{gargaro2023lettucemeta}. However, higher productivity can also require greater energy input and increase environmental impacts \citep{verteramochiu2024cea}. Efficient production therefore depends not only on maintaining suitable environmental conditions, but on adjusting those conditions according to crop demand as the plant develops.

Greenhouse sensors and imaging systems provide different views of this developing crop. Environmental sensors describe the climate surrounding the plant, while images can measure traits such as canopy size, leaf area, and structural development. Repeated observations can therefore show how the crop has grown over time. However, these measurements primarily describe the trajectory that the physical plant has already followed. They do not directly answer a more important management question: how would the same crop develop if its environment were changed now? For example, an observed growth trend alone cannot determine whether maintaining the current temperature and radiation will lead to the desired harvest state or whether another environmental trajectory would produce a better response. Addressing this question requires a model that begins from the observed crop state and predicts how the plant may develop under alternative future conditions.

Such prediction requires representing how environmental conditions influence plant physiology and structure. Leaves intercept radiation and assimilate carbon through photosynthesis. Part of this carbon is consumed through maintenance respiration, while the remaining carbon supports leaf expansion and the accumulation of structural dry matter. Increasing radiation can increase assimilation when light is limiting, while temperature affects both photosynthetic activity and respiratory demand \citep{sun2025lettuce}. As leaves expand, changes in their area, orientation, and overlap alter the radiation available for subsequent photosynthesis \citep{bailey2021resolution}. Outer leaves may remain exposed while younger leaves near the center of the rosette become increasingly shaded. Canopy development also affects air movement, creating differences between the ambient climate and the conditions at individual leaf surfaces that influence transpiration and photosynthesis \citep{vanwestreenen2020canopyclimate}. The environmental response of a plant therefore depends on both its physiological state and the arrangement of its leaves.

A crop digital twin provides a framework for connecting this modeled response with the development of a physical plant \citep{verdouw2021smartfarming}. Greenhouse sensor records can define a corresponding simulated environment, allowing physical and digital plants to develop under comparable conditions. Repeated measurements of the physical plant can then be compared with the digital plant's predicted biomass and structure. Differences between the two can guide examination of crop performance and model accuracy, informing either model updating or changes in crop management. Once this correspondence is established, alternative climate scenarios can be evaluated in the digital environment before they are applied to the physical crop.  Existing crop digital twin studies have progressed toward predictive crop management. \citet{ojo2026cropdt} combined plant images with environmental and nutrient records in a multimodal model, using a long short-term memory (LSTM) network to forecast lettuce fresh weight and leaf area alongside adaptive lighting control. These forecasts estimate plant traits but do not represent how individual leaves respond to the conditions they experience. \citet{frontzek2026adaptivedt} coupled a mechanistic lettuce model with parameter estimation and periodic biomass correction. Their case study tracked a single biomass state and tested correction with synthetic plant measurements. Neither demonstration established a physiological and structural reference for the development of a physical plant under corresponding greenhouse conditions. A difference between observed and predicted growth could reflect an inaccurate model or a physical crop whose response has changed. Without resolving this distinction, the comparison cannot reliably guide model correction or crop management.

Mechanistic growth models provide a physiological basis for building the digital plant needed for this comparison. Early lettuce models linked radiation, temperature, and carbon dioxide to dry matter accumulation \citep{vanhenten1994lettuce}. More recently, \citet{sun2025lettuce} developed and validated a lettuce growth model that relates air temperature, humidity, carbon dioxide, and shortwave radiation to photosynthesis, respiration, biomass accumulation, and leaf area development. However, its whole-plant representation does not resolve differences in environmental exposure and development among individual leaves.  Functional structural plant models
(FSPMs) provide a way to address this limitation by representing individual organs and their development
in three-dimensional space \citep{vos2010fspm}. Previous lettuce studies have separately modeled leaf appearance and expansion \citep{ko2026beta}, while ray-tracing approaches have shown how plant geometry can be used to resolve differences in radiation interception among leaves \citep{kim2020raytracing}. However, these components remain separate. The leaf model of \citet{ko2026beta} drives appearance and expansion from temperature and developmental stage rather than from the carbon assimilated by the plant, while the ray-tracing analysis of \citet{kim2020raytracing} calculated radiation on given plant structures without returning it to
growth. Thus, a whole-plant model can predict growth across climates but contains no individual leaves,
whereas leaf-level models can describe leaves and their light environment but lack the carbon balance that
grows them. To our knowledge, no existing lettuce model couples an established
whole-plant carbon balance to an organ-resolved three-dimensional structure in which ray-traced,
per-leaf radiation is returned to leaf physiology within the same simulation.

To address this gap, a unified functional–structural framework was developed to simulate lettuce growth using physiological information from individual leaves. Each leaf interacts with the local greenhouse microclimate, which influences its photosynthesis, transpiration, and expansion. Carbon assimilated by the leaves contributes to the whole-plant carbon balance, where maintenance respiration is first accounted for and the remaining carbon is distributed among leaves according to their age, size, and local conditions. The resulting leaf mass, area, and developmental age are then used to generate an evolving three-dimensional lettuce structure. As the plant develops, changes in leaf arrangement modify radiation interception and shading, which in turn influence subsequent leaf physiology and growth. In this way, the framework links local environmental conditions, leaf-level physiological processes, whole-plant carbon balance, and structural development. 

The framework was implemented in NVIDIA Isaac Sim to represent the spatial greenhouse environment. Greenhouse sensor records defined variations in air temperature, relative humidity, and carbon dioxide concentration, while ray tracing calculated the radiation intercepted by each leaf, including shading from surrounding leaves, neighboring plants, and greenhouse structures. The simulated plant could be driven by recorded greenhouse conditions or modified environmental scenarios while preserving the same physiological and structural relationships. This study develops and evaluates the resulting digital plant as a forward model for future integration with plant observations and greenhouse climate management. The primary objectives were to:

\begin{enumerate}
\item Develop a leaf-level lettuce model that links local environmental conditions to leaf physiology and whole-plant growth.
\item Generate an evolving three-dimensional lettuce structure from the physiological leaf states and integrate it into an NVIDIA Isaac Sim greenhouse, where radiation interception feeds back to leaf physiology and growth.
\item Evaluate the model against measured plants, using total dry weight to assess whole-plant growth and leaf number, largest leaf area, and canopy diameter to assess structural development.
\item Examine plant responses to environmental changes, spatial variation, and neighboring plants, together with the sensitivity of leaf allocation to local carbon supply and leaf-specific tipburn susceptibility.
\end{enumerate}

%% file: sections/2_1_overview.tex
\label{sec:modeling-framework-overview}

\begin{figure*}[!htbp]
    \centering
    \includegraphics[width=\textwidth]{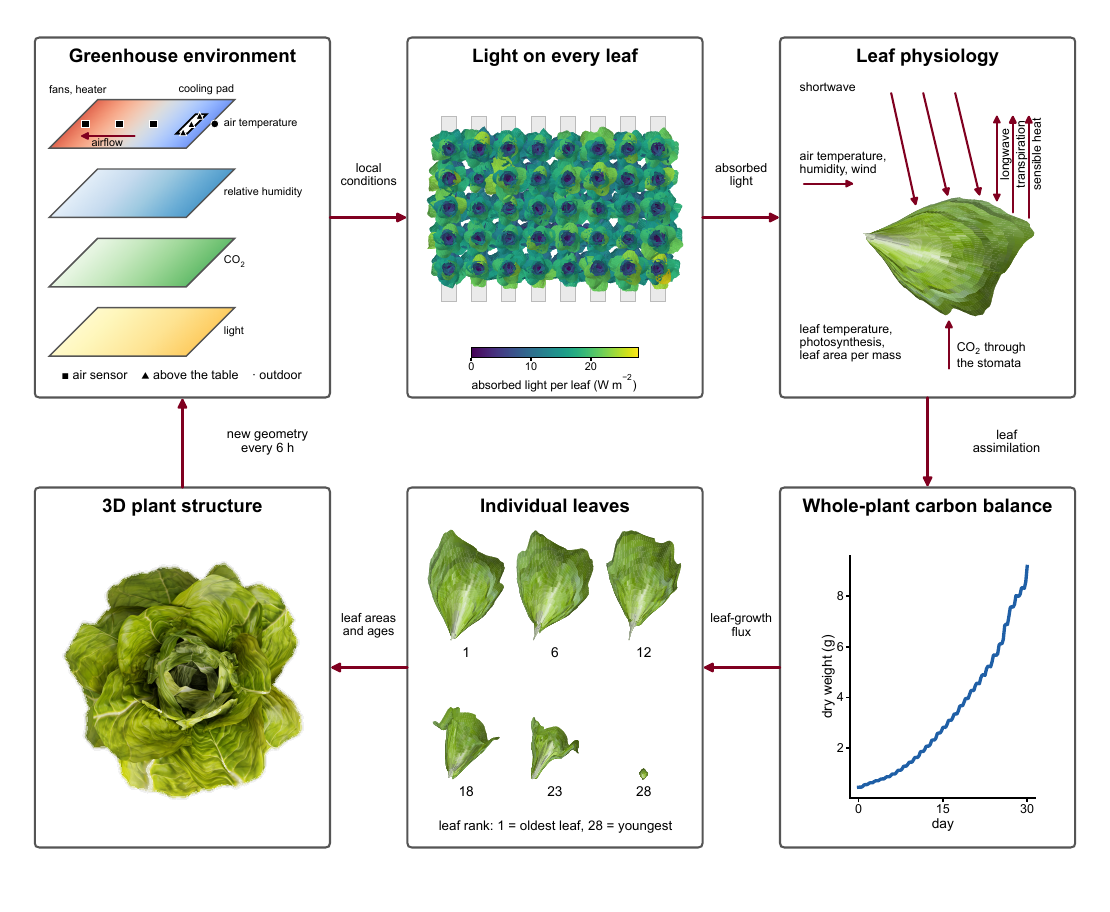}
    \caption{Overview of the modeling framework. The recorded greenhouse conditions and the current canopy give every leaf its own air conditions and absorbed light. Leaf physiology returns carbon gain and expansion, and the whole-plant carbon balance shares the resulting growth among the leaves. The updated leaves build the next canopy, which changes the light intercepted during the following interval.}
    \label{fig:framework}
\end{figure*}

Fig.~\ref{fig:framework} illustrated the modeling framework, which integrated four main components within NVIDIA Isaac Sim: (i) a leaf-level growth model, in which the carbon assimilated by the leaves entered a whole-plant carbon balance and the resulting growth was distributed among individual leaves, (ii) a three-dimensional plant generated from the predicted leaf areas and developmental ages, (iii) a simulated greenhouse environment, in which each leaf received the radiation and air conditions at its own position and these conditions were converted into leaf physiology, and (iv) the feedback through which the developing canopy altered radiation interception and subsequent growth. In the leaf-level growth model, the temperature, photosynthesis and expansion of each leaf were calculated from the radiation and air conditions at its position, the assimilated carbon entered the whole-plant carbon balance, and the growth obtained from that balance was distributed among the leaves according to their developmental age, size and local carbon supply. The predicted leaf areas and developmental ages then generated a three-dimensional plant in which every leaf surface kept the identity of its physiological leaf. In the simulated greenhouse, recorded conditions supplied the solar radiation, air temperature, relative humidity and carbon dioxide concentration, and ray tracing over the current canopy determined the radiation absorbed by each leaf after shading by its own plant and by neighboring plants. The absorbed radiation and the local air conditions were returned to the growth model for the following interval, so that the developing canopy altered the growth that followed. The following sections describe these four components in turn, followed by the numerical implementation and the greenhouse trials and simulations used to calibrate and evaluate the framework.

%% file: sections/2_2_1_plant_level.tex
\subsubsection{Whole-plant growth formulation}
\label{sec:plant_level_formulation}

The carbon-balance model of \citet{sun2025lettuce} provided the starting point for predicting lettuce dry matter. That model relates carbon gained through photosynthesis to carbon used for maintenance and growth. To represent the contribution of individual leaves, canopy assimilation was calculated from their exposed surfaces and returned to the same whole-plant balance. A leaf that received less light could therefore contribute less carbon to subsequent plant growth.

The whole-plant states were crop dry matter \(X_d\) (kg DM m\(^{-2}\) ground) and the carbohydrate-status regulator \(C_{\mathrm{buf}}\). Gross assimilation \(A_C\) from the leaves was converted to carbohydrate, from which maintenance respiration was deducted before net dry-matter growth was calculated. The regulator adjusted physiological activity according to carbohydrate availability and it was not counted as an additional pool of harvested dry matter. Leaf-level calculation of \(A_C\) is described in Section~\ref{sec:leaf_level_physiology}.

Net production was divided between shoot and root growth. The shoot fraction supplied the leaf-growth flux \(F_{\mathrm{leaf}}\), which was then shared among developing leaves. Thus, each leaf's exposure affected the carbon available to the whole plant, while the carbon balance determined how much new dry matter could be assigned to its leaves.

%% file: sections/2_3_leaf_dynamics.tex
\subsubsection{Leaf appearance and canopy development}
\label{sec:leaf_appearance_expansion_senescence}

Leaf number determined how many organs were available to receive growth and contribute to canopy development (Fig.~\ref{fig:plant-generation}A). New leaves were added using a continuous initiation counter \(N\). The rate of leaf initiation depended on temperature and carbohydrate availability. It increased from the base temperature to an optimum, then decreased as temperature approached the upper limit. The reference plastochron defined the time between successive leaf initiations under optimum temperature conditions. Cardinal temperatures were based on lettuce germination responses reported by \citet{cha2014lettuce}.

Leaf initiation was treated separately from leaf expansion because a newly initiated leaf does not immediately develop a large blade. \citet{ko2026beta} likewise treated romaine leaf appearance and expansion as separate processes. The initiation counter assigned each leaf an initiation time and developmental age. The developmental age of leaf \(i\) was expressed in plastochrons since its initiation, \(\tau_i=N-i\), and the leaf entered the growth allocation through an activation weight \(\phi_i\) that rose smoothly from zero at initiation to one once the leaf carried a blade. Once active, the leaf received dry matter through the allocation described in Section~\ref{sec:organ_resolved_reformulation}, and its \(SLA_i\) determined the blade area produced from that mass (Eq.~\eqref{eq:reference_allocation}). The resulting leaf ages, masses, and areas were then passed to the structural model to generate the next canopy geometry. Senescence and organ loss were not included in the simulations.

%% file: sections/2_2_2_organ_resolved.tex
\subsubsection{Individual-leaf allocation and growth}
\label{sec:organ_resolved_reformulation}

The plant-level carbon balance determined the total dry matter available for leaf growth, but it did not specify how that growth was divided among individual leaves. Each leaf was therefore tracked from initiation and assigned a share of the available growth according to its developmental age, current size, and local carbon supply. The activation weight \(\phi_i\) and developmental age \(\tau_i\) of Section~\ref{sec:leaf_appearance_expansion_senescence} determined when a leaf entered the allocation and its stage of development, and the relative sink demand \(s_i\) increased during early development and declined as the leaf matured, following a beta function \citep{marcelis1996sink,yan2004greenlab}.

Leaves of the same age could still receive different amounts of growth carbon because their current sizes could differ. To account for this, each leaf was compared with a moving reference area based on its rank within the canopy. The normalized potential-length profile \(B(r)\) for romaine lettuce \citep{ko2026beta} was converted into a relative area profile using the exponent \(\eta\):
\begin{equation}
\begin{aligned}
p_i&=\frac{\phi_i B(i/N)^\eta}{\sum_j\phi_j B(j/N)^\eta},\qquad
a_i^\star=\kappa\Big(\sum_j a_j\Big)p_i,\\
h_i&=\left[\max\left(1-\frac{a_i}{a_i^\star},0\right)\right]^\nu.
\end{aligned}
\label{eq:rank_area_constraint}
\end{equation}

Here, \(p_i\) gives the relative reference share for leaf \(i\), while \(a_i^\star\) is its moving reference area. The factor \(h_i\) reduces the growth priority of a leaf as its current area \(a_i\) approaches this reference. The parameters \(\kappa\), \(\eta\), and \(\nu\) control the overall reference-area scale, the conversion of the rank profile from length to area, and the strength of the size-based reduction, respectively. The total reference area is \(\kappa A\), where \(A=\sum_j a_j\) is the current total physiological leaf area. In this study, \(\kappa=1\). Because the reference area changed as the canopy grew, it did not prescribe a fixed final size for any leaf. Instead, it affected how the available growth was distributed among leaves at each time step. Total leaf area still depended on carbon gain, dry-matter allocation, and specific leaf area.

The developmental terms were then combined with the local carbon-supply modifier \(\psi_i\) to determine the final growth share of each leaf:
\begin{equation}
\begin{aligned}
w_i&=\frac{\phi_i s_i h_i\psi_i}{\sum_j\phi_j s_j h_j\psi_j},
& G_i&=\frac{F_{\mathrm{leaf}}}{\rho_c}w_i,\\
\frac{\mathrm dm_i}{\mathrm dt}&=G_i,
&\frac{\mathrm da_i}{\mathrm dt}&=SLA_iG_i.
\end{aligned}
\label{eq:reference_allocation}
\end{equation}

The normalized weight \(w_i\) determined the fraction of the available leaf-growth flux assigned to leaf \(i\). This gave the dry-matter growth rate \(G_i\), which increased the mass of that leaf, and specific leaf area \(SLA_i\) then converted the added dry matter into blade area. Because the weights were normalized, the total growth assigned across all leaves remained equal to the leaf growth supplied by the plant-level carbon balance. Local conditions influenced both \(\psi_i\) and \(SLA_i\), as described in Section~\ref{sec:leaf_level_physiology}.


\begin{figure*}[!htbp]
 \centering
 \includegraphics[width=\textwidth]{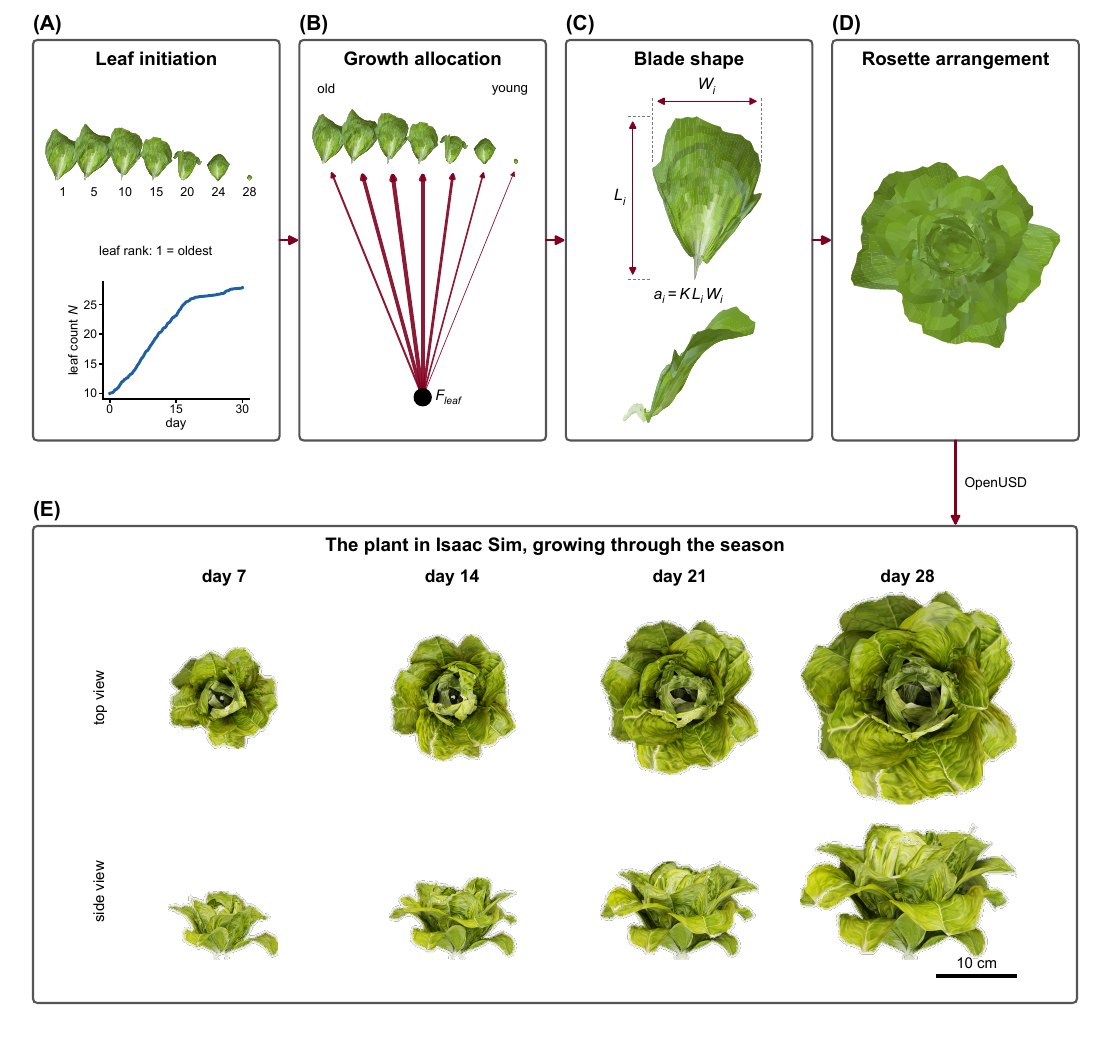}
 \caption{Generation of a lettuce plant from its individual leaves. (A) New leaves are initiated at a temperature-dependent rate and each leaf keeps its rank. (B) The leaf growth of the whole plant is shared among the leaves by age, size and local carbon supply, shown here at day 28. (C) The predicted area of a leaf sets its blade length and width before inclination and cupping are applied. (D) The blades are placed at the golden angle and the rosette is scaled to the target canopy diameter. (E) Top and side views of the same plant at 7, 14, 21 and 28 days after transplanting, drawn at one physical scale.}
 \label{fig:plant-generation}
\end{figure*}

%% file: sections/2_4_structure.tex
\label{sec:three_dimensional_structure_generation}

Leaf area alone does not describe how leaves are arranged or how they shade one another. The predicted area and developmental age of each leaf were used to generate a three-dimensional rosette, from which radiation interception was calculated. The rosette was generated in four steps: the dimensions of each blade were obtained from its area, the blade surface was formed along a curved midrib, the blades were placed around the crown in phyllotactic order, and the rosette was scaled to the canopy diameter expected from its total leaf area (Fig.~\ref{fig:plant-generation}C and D).

Blade length and width were derived from the physiological area of each leaf using a blade-shape coefficient and an age-dependent aspect ratio \citep{koyama2022shoot}:
\begin{equation}
a_i=K L_i W_i,\qquad W_i=r_i L_i,\qquad
r_i=r_{\mathrm{y}}+(r_{\mathrm{m}}-r_{\mathrm{y}})\,f_i.
\label{eq:blade_dimensions}
\end{equation}
In this equation, $L_i$ and $W_i$ denote the blade length and maximum width (m) of leaf $i$, respectively. The coefficient $K = 0.706$, obtained by integrating the normalized blade-width profile along the midrib, represents the ratio of blade area to the product of length and width. The parameter $r_i$ denotes the width-to-length ratio of leaf $i$, and $f_i$ its relative age, ranging from zero at leaf initiation to one for the oldest leaf. The width-to-length ratio declined from $r_y = 0.80$ in young leaves to $r_m = 0.68$ in the outermost leaves, indicating that mature outer leaves were proportionally narrower than younger leaves near the plant's heart.

The blade surface was constructed around a midrib of length $L_i$. Its orientation changed continuously from the insertion angle at the crown to the final bend at the leaf tip:
\begin{equation}
\begin{aligned}
\theta_i(\xi) &= \theta_{0,i} + \Delta\theta_i\, b(\xi),\\
\theta_{0,i} &= \theta_{y} + (\theta_{m}-\theta_{y})\, o_i,\\
\Delta\theta_i &= \Delta\theta_{y} + (\Delta\theta_{m}-\Delta\theta_{y})\, o_i.
\end{aligned}
\label{eq:midrib_pose}
\end{equation}
Here, $\xi$ is the normalized distance along the midrib, and $b(\xi)$ is a smooth bend profile increasing from zero at the base to one at the tip. The parameter $o_i$ represents the openness of leaf $i$, increasing from zero at initiation to one at maturity. As the leaf opened, its insertion angle decreased from $\theta_y = 82^\circ$ to $\theta_m = 44^\circ$, while the total midrib bend changed from $\Delta\theta_y = 58^\circ$ to $\Delta\theta_m = -78^\circ$. Young leaves therefore remained upright and curved inward around the heart, whereas mature leaves opened outward and drooped toward the canopy edge.

The blade surface was then formed around this midrib using the normalized width profile described above. For young, upright leaves, the blade margins were raised in proportion to leaf uprightness to reproduce the cupped shape characteristic of the heart. Margin ruffling and vein relief were progressively introduced as the leaves expanded. Each blade was discretized into 18 intervals along the midrib and 96 intervals across its width, yielding quadrilateral patches that formed the mesh used for the radiation calculations described in Section~\ref{sec:radiation_model}.

Individual blades were arranged around the crown in order of initiation, with successive leaves separated by the golden angle of $137.5^\circ$ and small random offsets in azimuth and insertion position. Older leaves occupied the outer canopy, while younger leaves remained closer to the crown. Leaves were placed sequentially from oldest to youngest; during placement, each new blade was lifted above the surface already occupied by older leaves at the corresponding azimuth and radial position, preventing blade intersections while allowing younger leaves to rest naturally above the existing canopy.

The overall rosette was scaled according to total physiological leaf area:
\begin{equation}
d^\star=\alpha_d A^{\beta_d},\qquad A=\sum_i a_i.
\label{eq:canopy_allometry}
\end{equation}
Here, \(A\) is total leaf area per plant (m\(^{2}\)), \(d^\star\) is the target canopy diameter (m), and \(\alpha_d\) and \(\beta_d\) are allometric parameters. Uniform scaling about the plant center preserved the relative arrangement of the leaves.

Because mesh deformation and scaling could change the geometric surface area, the mesh area was normalized before physiological fluxes were integrated:
\begin{equation}
\widetilde A_p=A_p^{\mathrm{mesh}}\frac{A}{A_{\mathrm{mesh}}},
\qquad A_{\mathrm{mesh}}=\sum_p A_p^{\mathrm{mesh}},
\qquad \sum_p\widetilde A_p=A.
\label{eq:patch_area_normalization}
\end{equation}
Here, \(A_p^{\mathrm{mesh}}\) is the geometric area of patch \(p\), \(A_{\mathrm{mesh}}\) is the total mesh area, and \(\widetilde A_p\) is the area used for physiological integration. This ensured that the total integrated leaf area matched the value predicted by the growth model, while the mesh itself continued to determine leaf position, orientation, visibility, and radiation interception.

Each leaf surface retained the identity of its corresponding physiological leaf, so radiation calculated on that surface was returned to the same leaf during the next growth update (Fig.~\ref{fig:plant-generation}E). Variation among simulated plants was introduced through persistent geometric differences and lognormal multipliers applied to initial size and selected physiological parameters \citep{baey2018mixed}.

%% file: sections/2_5_1_radiation.tex
\subsubsection{Radiation interception by the three-dimensional canopy}
\label{sec:radiation_model}

\begin{figure*}[!htbp]
    \centering
    \includegraphics[width=\textwidth]{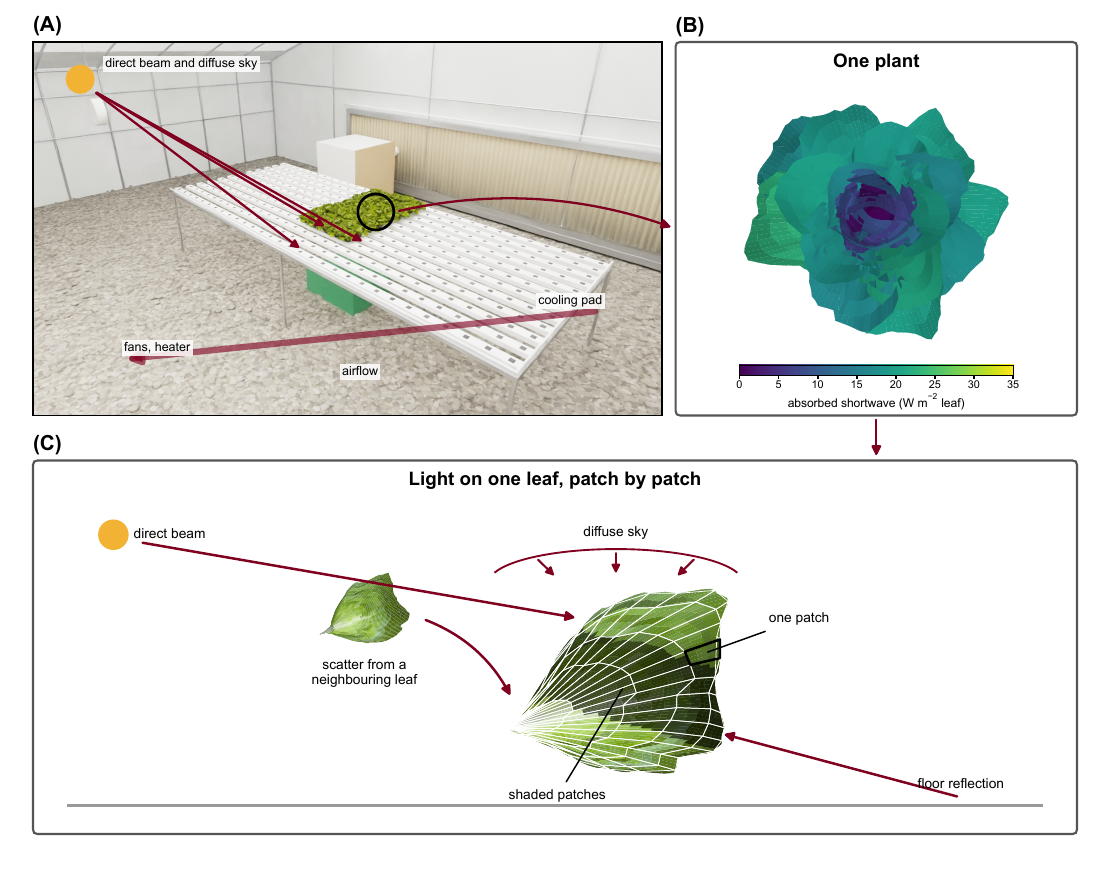}
    \caption{Radiation in the simulated greenhouse. (A) The NFT table in Isaac Sim at day 30, with the incident direct and diffuse light and the direction of the air field. (B) One plant colored by its mean absorbed shortwave radiation over days 23 to 30. The outer leaves receive more light than the enclosed center. (C) The light on the patches of one blade from the direct sun, 64 sky directions, floor reflection and scattering from neighboring leaves. Patches shaded by a neighbor are marked.}
    \label{fig:environment-exposure}
\end{figure*}

Radiation interception was recalculated as the canopy developed so that changes in leaf position and orientation could affect subsequent growth. Incident shortwave radiation was separated into direct and diffuse components using the clearness-index relationship of \citet{erbs1982diffuse}, while accounting for greenhouse glazing transmission and scattering. Direct radiation followed the solar direction, whereas diffuse radiation was sampled across the sky hemisphere (Fig.~\ref{fig:environment-exposure}C).

Each leaf surface was divided into patches to capture variation in radiation across the blade. Ray tracing determined which light directions were visible to each patch, accounting for shading from other leaves, neighboring plants, and greenhouse surfaces \citep{kim2020raytracing}. The absorbed radiation was then calculated from the incoming light, patch orientation, and leaf absorptance. Indirect radiation included first-order Lambertian reflection from the floor and Monte Carlo scattering with up to two reflections, while avoiding contributions already represented by the sky and floor components. Photosynthesis was calculated at the patch level before being combined for each leaf because the response to light is nonlinear \citep{bailey2021resolution}. An area-weighted shortwave radiation value was also calculated for each leaf and used in the leaf-temperature and expansion calculations. After each update of the plant geometry, radiation was traced again so that the next physiological calculation reflected the exposure of the newly developed canopy.

%% file: sections/2_5_2_microclimate.tex
\subsubsection{Spatial greenhouse environment}
\label{sec:greenhouse_microclimate_field}

The environmental conditions experienced by a leaf depended on its location within the greenhouse. Recorded measurements defined how conditions changed over time, while spatial variation was represented by interpolation between sensor locations for the simulation environment. To represent the spatial temperature distribution within the greenhouse, temperature was varied relative to a reference location following patterns reported for fan-and-pad greenhouses \citep{kittas2003gradients} (Fig.~\ref{fig:environment-exposure}A). Relative humidity was then calculated from the local temperature and a common vapor pressure using the Magnus relationship \citep{alduchov1996magnus}. Spatial variation in carbon dioxide was represented along the greenhouse and between canopy layers \citep{ito1970depletion}, while air speed was defined by the simulation scenario.

When measurements from multiple sensor locations were available, inverse-distance interpolation with vertical scaling was used to estimate conditions between sensors while preserving the recorded values at their locations. The resulting environmental variables were stored on a three-dimensional grid and sampled at leaf positions using trilinear interpolation. Leaf-level temperature and humidity were used in the energy-balance calculation, and humidity also influenced leaf expansion. Carbon dioxide for photosynthesis was evaluated at the plant position. The environmental field remained an imposed input to the model, while plant growth changed the canopy structure and its exposure to radiation.

%% file: sections/2_6_leaf_physiology.tex
\subsubsection{Leaf physiology and feedback to growth}
\label{sec:leaf_level_physiology}

The local environment influenced each leaf through three pathways: carbon assimilation, leaf expansion, and its share of the available plant growth. Leaf temperature was calculated first because it can differ from the surrounding air, especially under strong radiation. A two-sided energy balance combined absorbed shortwave radiation, longwave exchange, sensible heat, and evaporation to determine leaf temperature and transpiration \citep{campbellnorman1998biophysics}. Air speed and blade size affected boundary conductance, while absorbed radiation influenced stomatal conductance. The energy balance was solved using a linearized estimate followed by a nonlinear correction.

Photosynthesis was then calculated for each leaf patch using its absorbed photosynthetically active radiation and predicted leaf temperature. Diffusive and biochemical limits determined the maximum photosynthetic capacity, while patch-level light could also modify stomatal resistance. For positive LAI and photosynthetic capacity, the saturated gross assimilation capacity of patch \(p\) on leaf \(i\) was:
\begin{equation}
A_{\mathrm{sat},i,p}
=c_i\min(A_{\mathrm{diff},i,p},A_{\mathrm{bio},i})
+\frac{R_d}{c_\alpha LAI}.
\label{eq:patch_assimilation}
\end{equation}
Here, \(A_{\mathrm{diff},i,p}\) and \(A_{\mathrm{bio},i}\) are the diffusive and biochemical limits, respectively, in kg CO\(_2\) m\(^{-2}\) leaf s\(^{-1}\). The factor \(c_i\) represents acclimation to current or time-averaged leaf light and was normalized to an area-weighted mean of one. When acclimation was inactive, \(c_i=1\). Crop maintenance respiration \(R_d\) (kg CH\(_2\)O m\(^{-2}\) ground s\(^{-1}\)) was distributed across the active leaf area through LAI, and \(c_\alpha\) converted CO\(_2\) mass to carbohydrate mass.

The exponential light-response function then gave gross assimilation \(A_{L,i,p}\) for each patch. Patch fluxes were summed to obtain assimilation for each leaf and, subsequently, the canopy input to the whole-plant carbon balance:
\begin{equation}
\begin{aligned}
A_i&=\sum_{p\in i}\widetilde A_p A_{L,i,p},\qquad
\widetilde a_i=\sum_{p\in i}\widetilde A_p,\\
q_i&=\frac{A_i}{\widetilde a_i},\qquad
A_C=\rho_c\sum_i A_i.
\end{aligned}
\label{eq:leaf_to_plant_assimilation}
\end{equation}
Here, \(\widetilde A_p\) is the normalized area of patch \(p\), \(A_i\) is the total assimilation of leaf \(i\) (kg CO\(_2\) s\(^{-1}\)), and \(\widetilde a_i\) is its total integration area. The quantity \(q_i\) represents assimilation per unit leaf area, while planting density \(\rho_c\) converts the summed leaf fluxes to canopy assimilation \(A_C\). Calculating photosynthesis before combining the patch fluxes preserved the different responses of exposed and shaded regions of the same leaf (Fig.~\ref{fig:environment-exposure}B) \citep{long1993quantum,bailey2021resolution}.

The local environment also influenced how much blade area was produced from new dry matter. Leaf-specific area \(SLA_i\) was allowed to vary with absorbed light and relative humidity:
\begin{equation}
SLA_i=
\frac{SLA_{\mathrm{rf}}}
{\left[1+\beta_I(I_{\mathrm{ref}}-\overline Q_i)\right]
\left[1+\beta_{RH}(RH_{\mathrm{ref}}-RH_i)\right]}.
\label{eq:spatial_sla}
\end{equation}
Here, \(SLA_{\mathrm{rf}}\) is the reference specific leaf area, \(I_{\mathrm{ref}}\) and \(RH_{\mathrm{ref}}\) are reference light and relative humidity, and \(\beta_I\) and \(\beta_{RH}\) control the strength of these responses. Relative humidity was expressed as a fraction, and \(\overline Q_i\) represented either instantaneous or time-averaged absorbed light according to the simulation setting. This allowed leaves growing under different local conditions to produce different areas from the same dry-matter gain.

Finally, local carbon gain could modify how the common plant growth supply was distributed among leaves. A local-supply modifier \(\psi_i\) compared assimilation per unit area of leaf \(i\) with the sink-weighted mean across actively growing leaves:
\begin{equation}
\psi_i=(1-\lambda)+\lambda\frac{q_i+q_0}{\overline q+q_0},
\qquad
\overline q=\frac{\sum_{j\in\mathcal L}\phi_j s_jq_j}
{\sum_{j\in\mathcal L}\phi_j s_j}.
\label{eq:spatial_allocation}
\end{equation}
The parameter \(\lambda\), ranging from zero to one, controlled how strongly local assimilation influenced allocation. When \(\lambda=0\), allocation depended only on leaf development and size. The set \(\mathcal L\) contained leaves with valid assimilation and positive sink demand, while \(q_0>0\) prevented the ratio from becoming undefined under very low light. Leaves without a local assimilation estimate were assigned \(\psi_i=1\). This relationship was motivated by source-sink transport \citep{minchin2005phloem}, but it was not intended to represent phloem transport explicitly. Instead, \(\psi_i\) modified the allocation in Eq.~\eqref{eq:reference_allocation}. The resulting changes in leaf mass and area altered the canopy geometry and therefore the radiation environment used in the next growth step.

%% file: sections/2_7_tipburn.tex
\subsubsection{Leaf-specific tipburn susceptibility}
\label{sec:tipburn-risk-index}

Tipburn was examined because the youngest enclosed leaves can expand rapidly while receiving limited calcium through transpiration \citep{collier1983calcium}. An exploratory index compared each leaf's positive relative area-expansion rate with its transpiration latent-heat flux, using a small positive flux to keep the ratio defined. The resulting value, in m\(^{2}\) J\(^{-1}\), was higher when expansion demand was large relative to transpiration. Zero-area and non-expanding leaves were assigned zero demand. The index was used to compare leaves within the simulated canopy. It did not predict visible injury or represent calcium delivered by root pressure.

%% file: sections/2_9_implementation.tex
\label{sec:implementation_verification}

The growth model was coupled to the Isaac Sim greenhouse through OpenUSD, in which each leaf surface carried the identity of its physiological leaf, allowing plant physiology and three-dimensional structure to evolve together during simulation. Plant states were integrated using a fourth-order Runge-Kutta scheme with a 300-s time step. Every six hours, updated leaf states were used to regenerate plant geometry, after which radiation and local environmental exposure were recalculated for the next growth interval.

Radiation calculations and batched plant operations were GPU accelerated. Numerical settings used in the environmental-response simulations are summarized in Table~\ref{tab:numerical_configuration}. Radiation calculations for simple geometries were checked against analytical view factors, while scalar and GPU-batched implementations were compared under identical inputs to verify numerical consistency.

\begin{table}[t]
\centering
\caption{Numerical settings used in the environmental-response simulations.}
\label{tab:numerical_configuration}
\small
\begin{tabular}{p{0.42\linewidth}p{0.47\linewidth}}
\toprule
Quantity & Recorded setting \\
\midrule
Physiological integration & Fourth-order Runge-Kutta, 300 s \\
Geometry and exposure coupling & 21,600 s (6 h) \\
Spatial environmental grid & \(62\times32\times9\), trilinear interpolation \\
Diffuse-sky sampling & 64 directions \\
Scattering & Two bounces, 32 directions, seed 0 \\
Floor radiation treatment & Albedo 0.15 and spatial resolution 0.01 m \\
Local supply and expansion & \(\lambda=0.5\) and instantaneous absorbed-light SLA \\
Population and duration & Five simulated individuals over 30 days per scenario \\
Scalar/batch acceptance criterion & Relative state discrepancy \(\leq10^{-9}\) \\
\bottomrule
\end{tabular}
\end{table}

Because the normalization of Eq.~\eqref{eq:patch_area_normalization} preserved the total leaf area of each plant rather than the area of every individual mesh leaf, the plant-level correction \(c_A=A/A_{\mathrm{mesh}}\) and the leaf-level ratio \(\widetilde a_i/a_i\) were recorded during the simulations to identify cases in which geometric mismatch could affect leaf-level fluxes.

%% file: sections/plant_materials.tex
\subsubsection{Plant materials and greenhouse experiment}
\label{sec:plant_materials}

The plant observations were collected at the E.W. Shell Fisheries Center of Auburn University in Auburn, Alabama, USA. Lettuce (\textit{Lactuca sativa} L.) representing leaf, romaine, butterhead and Latin types was grown in a temperature-controlled glasshouse section of approximately 27~m by 9~m. The measurements used for this model were from breeding line 10207. Seeds were raised for 14 days in a container farm under artificial light and automated irrigation. Uniform seedlings were then transplanted into a hydroponic nutrient film technique (NFT) system. Plants were spaced approximately 17.8~cm within and between channels. The recirculating nutrient solution was maintained at an electrical conductivity of 1.8~dS~m$^{-1}$ and pH 5.8. A completely randomized design was used with 16 plants per cultivar.

The plants received natural sunlight without supplemental lighting. Two heaters, one gable fan, two slant fans and an evaporative cooling pad regulated greenhouse temperature. The heater setpoints were 17.7 to 18.3~$^{\circ}$C during the day and 12.2 to 12.7~$^{\circ}$C at night. Fans and cooling pads were activated at 26.1 to 27.7~$^{\circ}$C during the day and 20.5 to 22.2~$^{\circ}$C at night, while small circulation fans moved air around the plants. Wireless sensors along the greenhouse and above the NFT table recorded air temperature, relative humidity and carbon dioxide throughout the growing period.

Seedling fresh and dry weight, leaf area, leaf number and plant diameter were measured at transplanting to define the initial state. The day of transplanting was designated as 0 days after transplanting (DAT). Additional plants were sampled destructively at 7, 14, 21 and 28 DAT. Shoots and roots were separated and weighed fresh, then dried to obtain shoot, root and total dry weight. Leaf area index (LAI), leaf number and the widest canopy diameter were recorded, and each plant was assessed for tipburn. Different plants of the same cultivar were therefore measured at successive dates. Plant observations from one Auburn trial were used for calibration, while a separate trial was reserved for evaluation.

Total dry weight was the main measure of whole-plant growth, while LAI, leaf number and canopy diameter tested different parts of structural development. Model dry weight in g plant\(^{-1}\) was calculated as \(1000X_d/\rho_c\) and compared with measured shoot plus root dry weight. The largest modeled leaf was the maximum physiological area \(a_i\) at each sampling, so its identity could change over time. Modeled leaf number was the initiation counter \(N\), and canopy diameter was the target \(d^\star\) of Eq.~\eqref{eq:canopy_allometry}.

After calibration, the model was driven by the environmental record and initial state of the evaluation trial without refitting parameters. Predicted and measured traits were compared at 7, 14, 21 and 28 DAT without shifting the time axis. Each measured mean represented different plants because sampling was destructive. The correlation was summarized by root mean square error (RMSE), mean signed error (simulation minus measurement) and relative RMSE (RRMSE), calculated as RMSE divided by the mean measured value and multiplied by 100.

%% file: sections/2_8_calibration.tex
\subsubsection{Model parameters and environmental inputs}
\label{sec:model_calibration_evaluation}

The developmental and structural relationships added to the starting carbon-balance model required initial parameter values. Plant observations from one Auburn greenhouse trial were used to estimate parameters requiring calibration. Those values were fixed before evaluation with a separate Auburn trial, whose plant measurements were not used during fitting. Table~\ref{tab:recorded_parameters} distinguishes estimated parameters from fixed settings.

\begin{table*}[!tbp]
\centering
\caption{Development and allocation parameters.}
\label{tab:recorded_parameters}
\small
\begin{tabular}{@{}p{0.10\linewidth}p{0.40\linewidth}p{0.14\linewidth}p{0.26\linewidth}@{}}
\toprule
Parameter & Meaning and unit & Value & Source \\
\midrule
\(\kappa\) & Rank reference scale (1) & 1.0 & Structural assumption \\
\(\eta\) & Rank-profile area exponent (1) & 1.0 & Structural assumption \\
\(P_{\mathrm{ref}}\) & Reference plastochron (s leaf\(^{-1}\)) & 53923.3~s & Auburn calibration trial \\
\(\lambda\) & Local-supply blend strength (1) & 0.5 & Selected setting \\
\(q_0\) & Assimilation regularizer (kg CO\(_2\) m\(^{-2}\) leaf s\(^{-1}\)) & \(10^{-12}\) & Numerical floor \\
\(\rho_c\) & Planting density (plants m\(^{-2}\)) & 31.6 & Auburn trial layout \\
\bottomrule
\end{tabular}
\end{table*}

The environmental records of each trial were collected by a wireless sensor network installed in the greenhouse (Fig.~\ref{fig:framework}). Seven LoRaWAN sensors (IOT-S500CO2, Linovision) measured air temperature, relative humidity and carbon dioxide concentration. One was placed near the exhaust fans and one near the middle of the house to record the longitudinal gradient, and five were placed on and above the NFT table in three levels, three at crop height and one at each of two heights above the canopy, to record the vertical profile over the crop. A separate quantum sensor at the top of the NFT table measured photosynthetically active radiation (PAR) and supplied the incident-radiation series. The sensors transmitted to a LoRaWAN gateway (IOT-G63, Linovision) with an embedded network server, which published the readings over MQTT to a local server, where they were decoded and stored at 5 min intervals. These records supplied the temporal boundary conditions of the simulations, and the sensor positions defined the interpolation of Section~\ref{sec:greenhouse_microclimate_field}.

For each Auburn trial, recorded air temperature, relative humidity and carbon dioxide were applied as environmental inputs to the model in Isaac Sim. The PAR record supplied the light entering the simulated greenhouse, and ray tracing then calculated the light absorbed by individual leaves. The simulation began at transplanting, using that trial's seedling measurements as its initial plant state.

\subsubsection{Simulation experiments}
\label{sec:environmental_scenarios}

Simulation experiments were conducted to examine the response of the model to changes in climate, plant position, neighboring plants and local carbon allocation, and to evaluate the tipburn index. In each comparison, the initial plant states and model parameters were held fixed and only the tested condition was changed. The scenario definitions, the diagnostic evaluation of the prescribed air field, the tipburn comparison and the allocation ablation are reported in full in \ref{app:supporting}, and the main text reports the outcomes.

\textit{Environmental responses.} Five plants were simulated for 30 days under the reference record and under five sustained changes: air temperature lowered and raised by 2~\(^{\circ}\mathrm{C}\), radiation multiplied by 0.7 and by 1.3, and daytime carbon dioxide concentration raised by 200~ppm. Gross assimilation, maintenance respiration, retained assimilation and final dry weight were compared with the reference run.

\textit{Spatial scenarios.} Forty plants were simulated in an eight-channel by five-hole NFT block. Five scenarios separated the effects of the prescribed air field, shading by neighboring plants and plant-to-plant variability (Table~\ref{tab:spatial-scenarios}), and each was compared with a uniform-air, identical-plant reference.

\textit{Tipburn susceptibility.} The youngest one-third of the leaves larger than 1~cm\(^2\) defined the enclosed group, and their mean index gave a plant-level susceptibility index. The simulated index was compared with the observed tipburn severity and plant size.

\textit{Allocation sensitivity.} Paired simulations with \(\lambda=0.5\) and \(\lambda=0\) were run under the same initial and environmental conditions, and final dry weight, largest-leaf area, total leaf area and individual leaf areas were compared between the two settings.

%% file: sections/results_1_3.tex
\subsection{Simulated and measured plant growth}
\label{sec:result_dry_weight}

\begin{figure*}[htbp]
 \centering
 \includegraphics[width=\textwidth]{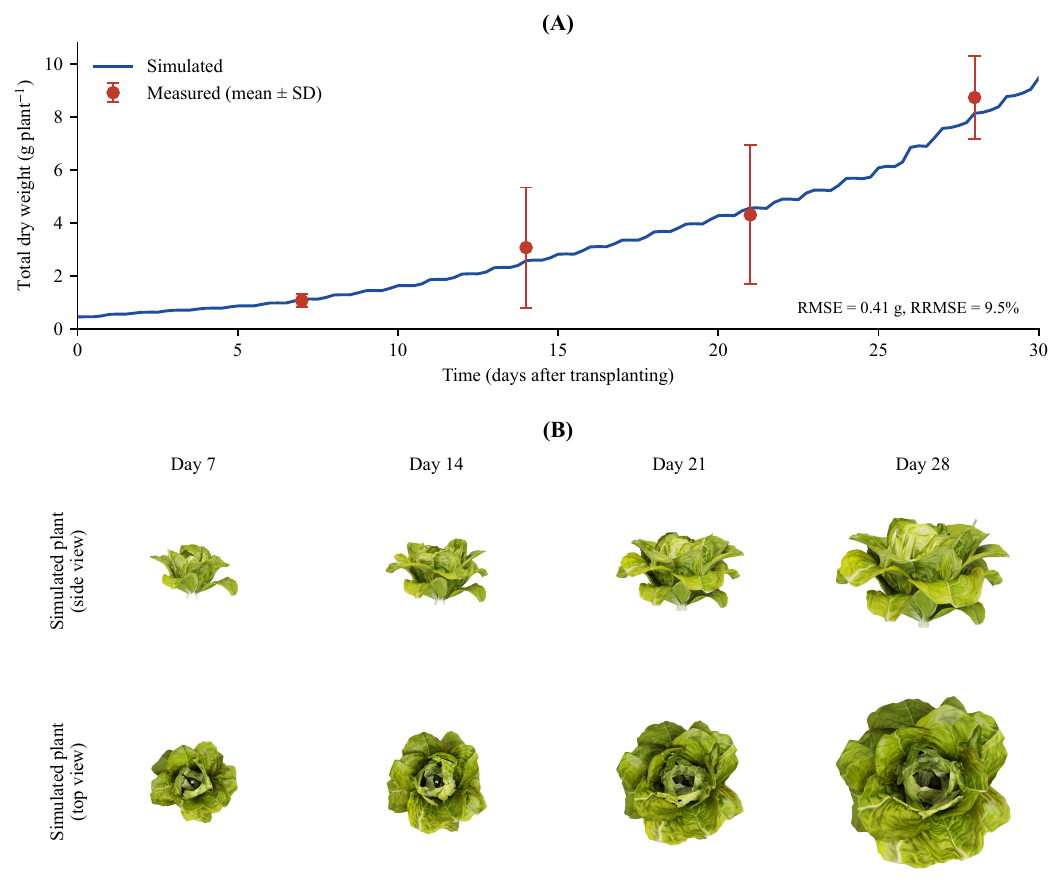}
 \caption{Simulated and measured total dry weight. (A) Simulated total dry weight from transplanting to day 30 and measured means at days 7, 14, 21 and 28, with error bars showing one standard deviation among the measured plants. (B) Side and top views of the simulated plant at the four sampling times, drawn at one physical scale.}
 \label{fig:result1-growth}
\end{figure*}

Fig.~\ref{fig:result1-growth} shows the simulated growth of lettuce during the 30-day period and the measured dry weight at the four weekly sampling dates. The simulated total dry weight increased from 1.12~g plant\(^{-1}\) on day 7 to 8.13~g plant\(^{-1}\) on day 28, while the measured mean increased from 1.07 to 8.73~g plant\(^{-1}\) over the same period (Fig.~\ref{fig:result1-growth}A). The simulation tracked the measurements throughout the experiment, with an RMSE of 0.41~g plant\(^{-1}\) and an RRMSE of 9.5\%. The RRMSE expresses the RMSE as a percentage of the mean of the four measured values. The mean signed error was \(-0.19\)~g plant\(^{-1}\), and the simulated values at all four sampling dates were within one standard deviation of the corresponding measurements. The largest difference occurred on day 28, when the model underestimated the measured mean by 0.59~g plant\(^{-1}\), or 6.8\%. Both the measured and simulated plants showed an accelerating pattern of dry-matter accumulation as development progressed. This behavior follows from the feedback in the model between leaf expansion and carbon acquisition: carbon allocated to expanding leaves increased the photosynthetic surface of the plant, which increased radiation interception and the carbon available for further growth. Similar accelerating growth before canopy closure has been reported for lettuce \citep{tei1996groundcover} and is the basis of dynamic lettuce growth models \citep{vanhenten1994lettuce}. The correspondence between the two trajectories therefore suggests that the coupling among radiation interception, carbon assimilation and leaf-level dry-matter allocation reproduced the rate of biomass accumulation. The alternating sign of the residuals, with small overestimation at some sampling dates and underestimation at others, indicates no strong systematic bias, although the growing underestimation toward the final sampling shows that the discrepancy increased as the canopy developed.

Fig.~\ref{fig:result1-growth}B illustrates the structural development predicted by the model. The side and top views show a progressive increase in plant size from day 7 to day 28, with continued expansion of the outer leaves and development of the central rosette. The increase in projected canopy area is evident in the top views, and the side views show the growing vertical and lateral extent of the foliage. These geometries were not prescribed but generated from the predicted area and developmental state of the individual leaves, so changes in plant structure fed back into radiation interception and carbon assimilation. This feedback matters most later in development, when overlapping leaves and a denser rosette increase self-shading, and differences between the simulated and actual canopy at that stage may contribute to the larger dry-weight deviation at day 28.

The prediction accuracy also compares well with previously reported lettuce growth models. \citet{sun2025lettuce} reported RRMSE values of 10.5 to 24.9\% for dry-weight prediction across three greenhouse climates, whereas the present leaf-level formulation reached 9.5\% under the evaluation conditions used here. The two figures come from different greenhouses, cultivars and sampling schemes, so they indicate a comparable order of accuracy rather than an improvement attributable to leaf resolution. Lower errors, with mean absolute percentage errors of 2.8 to 8.4\%, have been reported for nonlinear growth curves fitted directly to measured lettuce canopy data \citep{li2022morphological}, but such curves describe the measurements they are fitted to rather than predicting growth from the environment and an initial plant state. The present model generated the biomass trajectory from the radiation received by individual leaves, their carbon assimilation and the allocation of dry matter. The results therefore indicate that resolving light and air conditions at the leaf level reproduced the measured growth trajectory at an accuracy comparable to established plant-level formulations, while additionally carrying the organ-level state that those formulations do not represent.

\subsection{Simulated and measured structural development}
\label{sec:result_structure}

\begin{figure*}[htbp]
 \centering
 \includegraphics[width=\textwidth]{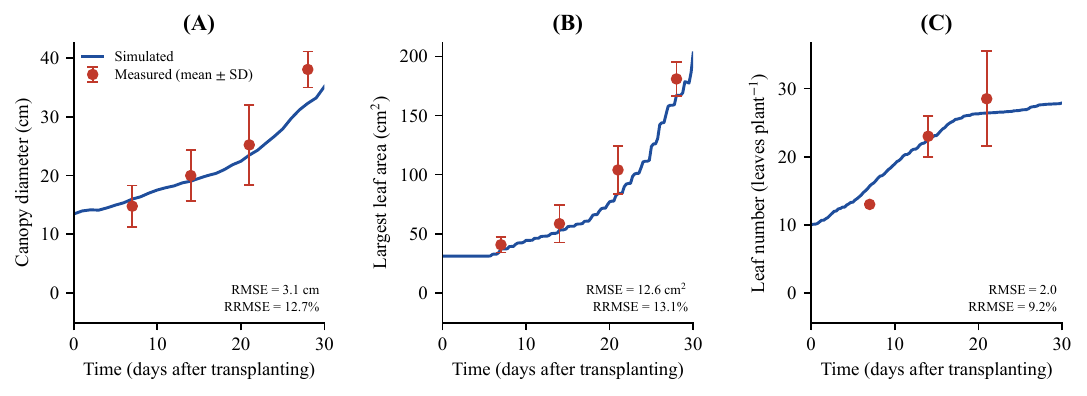}
 \caption{Simulated and measured structural development at the weekly samplings. (A) Canopy diameter. (B) Largest single-leaf area. (C) Leaf number. Lines are the simulated plant, symbols are measured means, and error bars show one standard deviation where at least two measurements were available. RMSE and RRMSE are given in each panel.}
 \label{fig:result2-structure}
\end{figure*}

Fig.~\ref{fig:result2-structure} compares the simulated canopy diameter, largest-leaf area and leaf number with the measurements at the weekly samplings. Canopy diameter increased from 16.0~cm on day 7 to 32.3~cm on day 28, compared with measured means of 14.8 and 38.1~cm (Fig.~\ref{fig:result2-structure}A). The RMSE was 3.11~cm, the RRMSE 12.7\% and the mean signed error \(-1.82\)~cm. The simulated diameter was slightly above the measured mean on day 7 and below it at the three later samplings, with the largest difference of 5.77~cm on day 28. This indicates that the simulated rosette became more compact than the measured plants during later development while its biomass stayed close to the measurements. In the model, diameter follows from the length of the blades and from the fixed inclination and placement rules of Section~\ref{sec:functional_structural_model}, so a growing shortfall in diameter would arise if the outer blades expanded less than the measured ones or if the fixed inclination held them more upright than the real leaves. Canopy spread in lettuce depends on both blade size and leaf angle, and the diameter measurements cannot separate the two. The trend is, however, consistent with the dry-weight deficit at day 28 in Section~\ref{sec:result_dry_weight}, since a narrower rosette intercepts less light.

The LAI recorded in the trial was the area of the largest blade of each plant, so it was compared with the largest simulated leaf, \(\max_i a_i\), for consistency (Fig.~\ref{fig:result2-structure}B). The simulated largest leaf grew from 37.0 to 167.2~cm\(^2\) and the measured mean from 40.8 to 180.9~cm\(^2\), with an RMSE of 12.59~cm\(^2\), an RRMSE of 13.1\% and a mean signed error of \(-10.75\)~cm\(^2\). The largest underestimation occurred on day 21, 83.9 against 104.0~cm\(^2\). In the model, the area of a leaf is the dry matter it received multiplied by the area produced per unit dry matter, and its share of dry matter follows the developmental sink demand and the rank reference of Section~\ref{sec:organ_resolved_reformulation} \citep{marcelis1996sink}. The underestimation of the largest blade from day 21 onward therefore points to the allocation or to the specific leaf area rather than to the carbon supply, which Section~\ref{sec:result_dry_weight} showed to be close to the measurements, and it is consistent with the diameter shortfall because a smaller largest blade produces a narrower rosette.

Leaf number had the lowest relative error of the three traits, with an RMSE of 1.99 leaves plant\(^{-1}\) and an RRMSE of 9.2\% over the three dates with counts (Fig.~\ref{fig:result2-structure}C). The model predicted 15.7 leaves on day 7 against the single observation of 13, and 22.6 and 26.4 leaves on days 14 and 21 against measured means of 23.0 and 28.5, so the initiation rate was slightly high in the first week and slightly low afterward, and the mean signed error of 0.05 leaves reflects this cancellation rather than a uniform fit. Leaf appearance in the model advances with temperature through cardinal temperatures taken from lettuce germination \citep{cha2014lettuce}, and the counter is continuous, so fractional values mark progress toward the next leaf and the observed counts depend on the size at which a leaf was recorded.

The structural accuracy is of the same order as models dedicated to leaf development. The leaf appearance module of \citet{ko2026beta}, calibrated on 437 romaine plants, predicted leaf number with an RMSE of 4.26 leaves in calibration and 2.24 leaves in independent evaluation. The length to area allometry used there for individual leaves carried an RMSE of 14.6~cm\(^2\), whereas the present model reached 1.99 leaves and 12.59~cm\(^2\) for the largest blade. The two error figures rest on very different sample sizes, so they indicate a comparable order of accuracy rather than a ranking. Nonlinear growth curves fitted to lettuce canopy measurements described canopy area with mean absolute percentage errors of 2.8 to 8.4\% \citep{li2022morphological}, and the 12.7\% for canopy diameter, predicted from the environment and an initial plant state, is of the same order. The model therefore produced approximately the right number and size of leaves, while the spread of the rosette late in the season was underestimated. Leaf appearance and leaf expansion respond differently to the environment \citep{ko2026beta}, and individual leaf area and leaf number contribute separately to canopy area in lettuce \citep{kong2021leafnumber}, so resolving the leaves individually is what allowed these traits to be evaluated one by one.

\subsection{Plant response to sustained environmental changes}
\label{sec:result_environment}

\begin{figure*}[htbp]
 \centering
 \includegraphics[width=\textwidth]{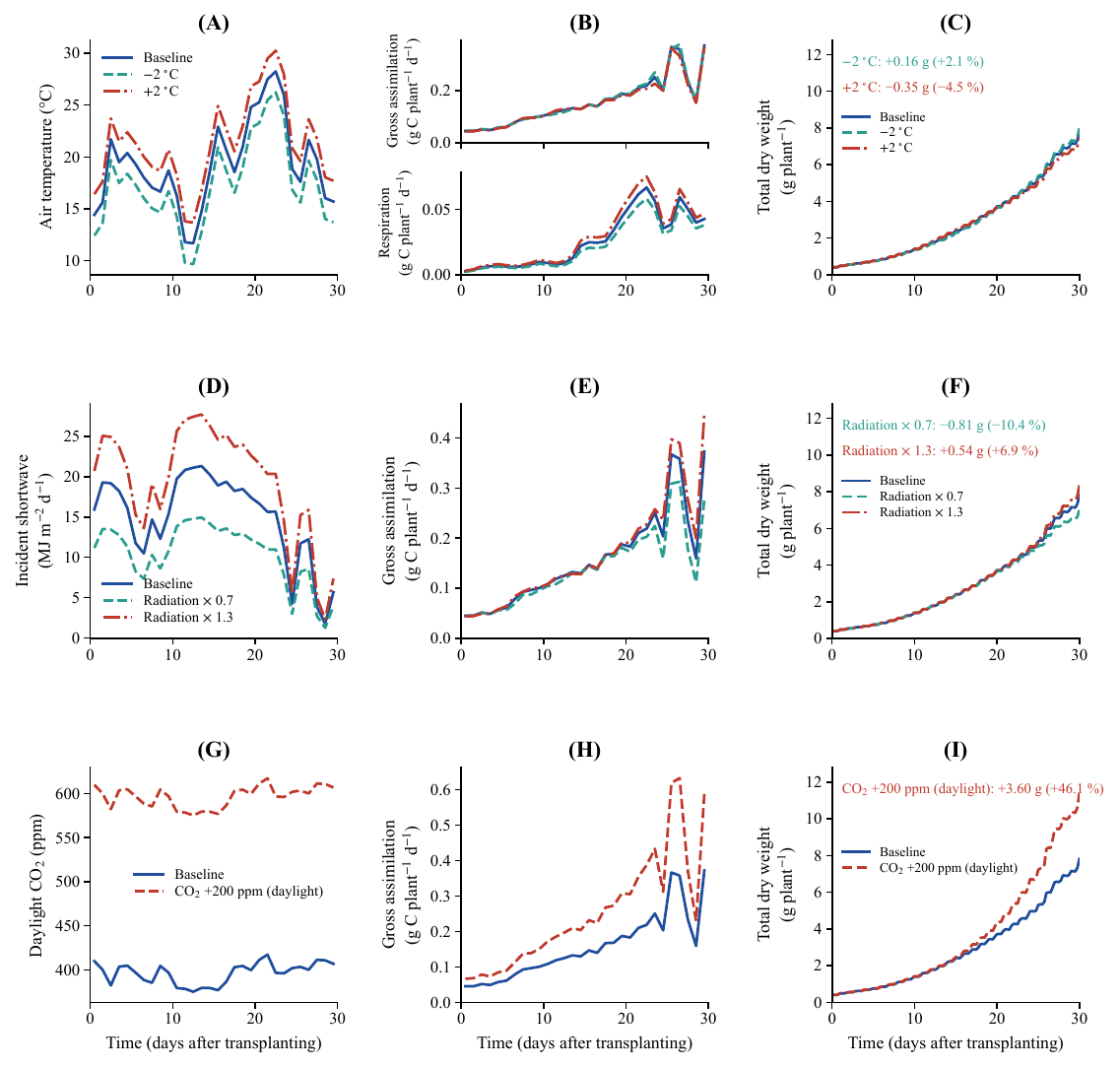}
 \caption{Simulated response to sustained environmental changes over a 30-day season. Rows are air temperature (A to C), incident shortwave radiation (D to F) and daylight carbon dioxide concentration (G to I). The left column is the applied forcing, the middle column the physiological response, and the right column the accumulated total dry weight. Curves are means of the same five simulated plants. Labels give the difference from the reference day-30 mean in grams per plant and in percent.}
 \label{fig:result3-environment}
\end{figure*}

To examine the influence of air temperature, incident shortwave radiation and carbon dioxide concentration on lettuce growth, the same five plants were simulated under the reference record and under sustained changes in each factor (Fig.~\ref{fig:result3-environment}). Positions, initial states and parameters were held constant. Under the reference record, with a mean air temperature of 19.3~\(^{\circ}\mathrm{C}\), 14.9~MJ~m\(^{-2}\)~d\(^{-1}\) of shortwave radiation, 396~ppm daylight carbon dioxide and 52\% relative humidity, mean total dry weight reached 7.81~g plant\(^{-1}\) on day 30. Temperature produced the smallest dry-weight response within the tested range (Fig.~\ref{fig:result3-environment}A to C). Lowering air temperature by 2~\(^{\circ}\mathrm{C}\) increased seasonal gross assimilation by 0.8\% and reduced maintenance respiration by 13.4\%, which raised day-30 dry weight by 0.16~g plant\(^{-1}\) (+2.1\%), whereas raising it by 2~\(^{\circ}\mathrm{C}\) reduced assimilation by 2.6\%, increased respiration by 13.0\% and lowered dry weight by 0.35~g plant\(^{-1}\) (\(-4.5\)\%). The larger proportional change in respiration shows that the response came mainly from the carbon cost of maintenance, so around the reference record warmer air raised the expenditure faster than the income. Relative humidity was held at its recorded value, so the vapor pressure deficit changed with temperature, and two offsets around one record cannot identify an optimum growing temperature. Radiation produced larger differences (Fig.~\ref{fig:result3-environment}D to F). Reducing incident radiation to 0.7 times the reference decreased gross assimilation by 10.5\% and dry weight by 0.81~g plant\(^{-1}\) (\(-10.4\)\%), while increasing it to 1.3 times raised assimilation by 6.3\% and dry weight by 0.54~g plant\(^{-1}\) (+6.9\%). The loss under shading was 1.5 times the gain under the corresponding increase because the light response of a leaf saturates: additional light on a surface near saturation adds little, whereas removing light from that surface takes away assimilation on the steep part of the response \citep{long1993quantum}. Because photosynthesis was evaluated patch by patch before aggregation, this nonlinearity acted separately on the exposed and shaded parts of the canopy \citep{bailey2021resolution}, and each treatment also grew its own canopy, so the day-30 difference includes the leaf area that the altered carbon gain built during the season.

Carbon dioxide enrichment produced the largest response (Fig.~\ref{fig:result3-environment}G to I). Adding 200~ppm to the daylight concentration increased seasonal gross assimilation by 61.1\% and day-30 dry weight from 7.81 to 11.42~g plant\(^{-1}\) (+46.1\%). The additional carbon changed development as well: mean initiated leaf number rose from 28.5 to 35.2 and total leaf area from 3288 to 5104~cm\(^2\) plant\(^{-1}\), so the enriched plants were both heavier and structurally larger. Biomass increased less than gross assimilation because maintenance respiration rose by 25.0\% with plant size and because the fraction of assimilation retained after carbon-buffer regulation fell from 96.0 to 85.7\%. The response therefore depended on the balance among carbon acquisition, regulation and maintenance over the season, not on the concentration change alone.

\begin{figure*}[htbp]
 \centering
 \includegraphics[width=0.92\textwidth]{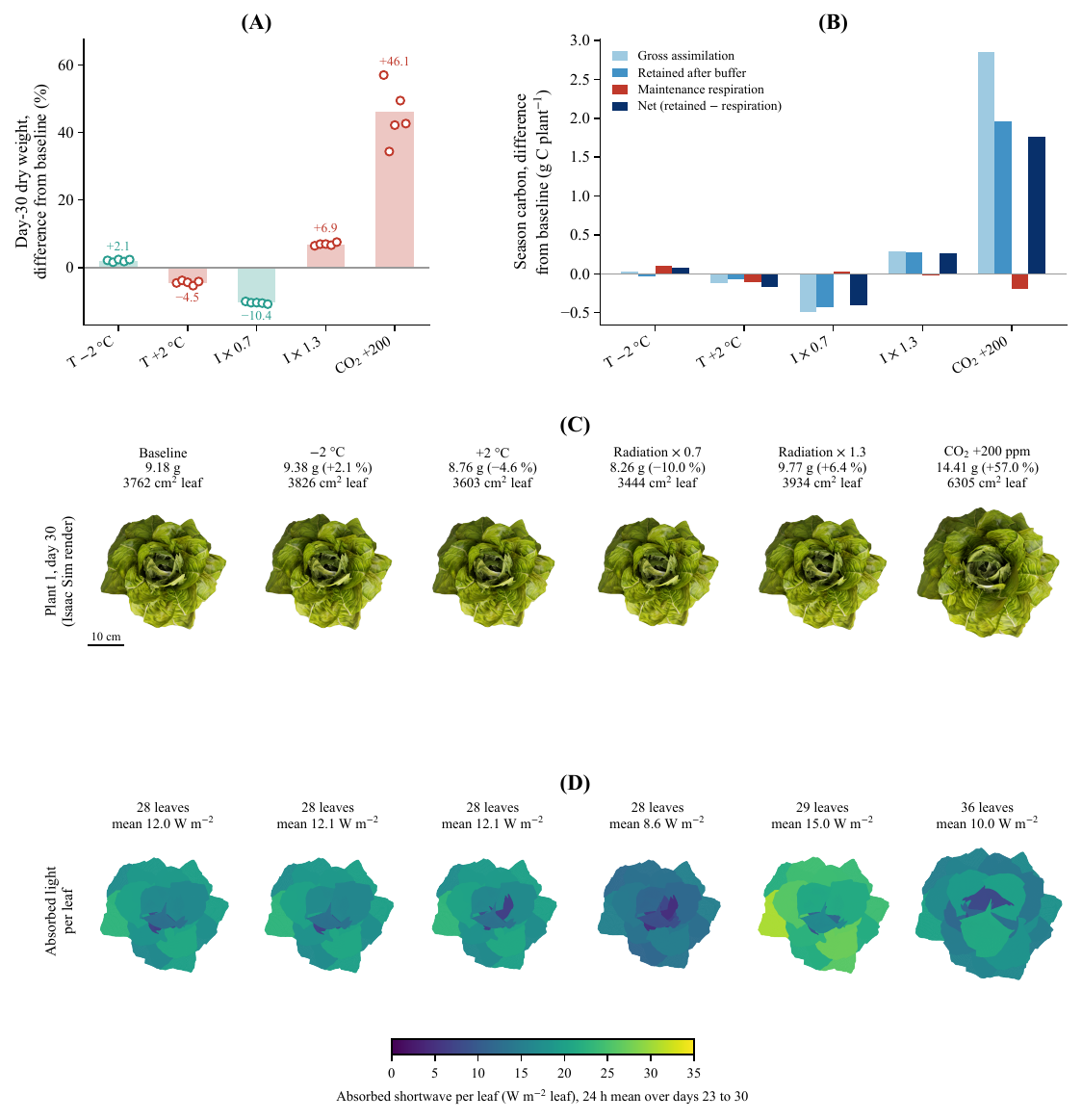}
 \caption{Dry-weight, carbon-budget and structural responses to the environmental changes. (A) Day-30 dry-weight difference from the reference for each of the five plants, with bars for the scenario means. (B) Difference from the reference seasonal carbon budget: gross assimilation, assimilation retained after carbon-buffer regulation, maintenance respiration and net retained carbon. (C) Top views of one plant at day 30 under each scenario at one physical scale, labeled with its dry weight, its difference from the reference and its total leaf area. (D) Absorbed shortwave radiation of the same plant by leaf, averaged over days 23 to 30.}
 \label{fig:result3-plants}
\end{figure*}

Fig.~\ref{fig:result3-plants} shows how these responses were expressed among the plants and within the canopy. The direction of each dry-weight response was the same for all five individuals (Fig.~\ref{fig:result3-plants}A), and retained assimilation minus maintenance respiration tracked the relative dry-weight change to within 1.7 percentage points across the scenarios (Fig.~\ref{fig:result3-plants}B), which supports the carbon-budget reading of the responses, although the budget and the dry weight are outputs of the same model. The rendered plants and their per-leaf radiation maps (Fig.~\ref{fig:result3-plants}C and D) show the responses as rosettes of different size, with the lowest absorbed radiation always on the enclosed inner leaves. One incident-radiation value therefore did not describe the exposure of every leaf, and the response of the plant was built from the light each leaf received \citep{vos2010fspm,vanwestreenen2020canopyclimate}.

The temperature and radiation responses also compare well with measurements once the growing conditions are taken into account. Growth-chamber experiments with constant setpoints report the opposite temperature sign over a wider range. \citet{tarr2025temperature} found the fresh mass of butterhead lettuce increasing by 18\% from a mean daily temperature of 20 to 26~\(^{\circ}\mathrm{C}\) under 300~\(\mu\)mol~m\(^{-2}\)~s\(^{-1}\) and 500 to 1200~\(\mu\)mol~mol\(^{-1}\) carbon dioxide, conditions under which warming raises assimilation faster than respiration, whereas under the recorded greenhouse radiation the modeled assimilation gain was small. The sign of the temperature response therefore depends on the light and carbon dioxide available. For radiation, lowering the daily light integral to 55\% for the last 12 days of indoor production reduced lettuce yield by up to 22 to 26\% under light-limited artificial lighting \citep{ertle2023dailylight}. The smaller \(-10.4\)\% here for a 30\% reduction over the whole season is consistent with the recorded 14.9~MJ~m\(^{-2}\)~d\(^{-1}\) placing the exposed leaves closer to saturation than the 200 to 300~\(\mu\)mol~m\(^{-2}\)~s\(^{-1}\) of those chambers.

Measured responses to enrichment are smaller. In a greenhouse nutrient film system, supplementing to 800~ppm increased lettuce dry weight by 21.4\% \citep{singh2020co2}, and a meta-analysis of 107 vegetable studies gave a mean yield gain of 34\% at 827~\(\mu\)mol~mol\(^{-1}\) \citep{dong2020eco2}, with the largest gains between 400 and 800~ppm and little beyond \citep{tarr2025temperature}. The 46.1\% increase for a 200~ppm addition is above that range, and photosynthetic acclimation and nutrient limitation, which the model does not represent, are the processes that would move it toward the 21 to 34\% reported for greenhouse and plant-factory lettuce. The model therefore responded to temperature, radiation and carbon dioxide within the range measured for lettuce, with the size of each response set by the light and carbon that each leaf received.

%% file: sections/results_4_6.tex
\subsection{Growth and leaf exposure in a planted block}
\label{sec:result_spatial}

To understand how the position of a plant within a planted block and the shading by its neighbors change its growth, 40 plants were simulated on the NFT table under the five scenarios of Table~\ref{tab:spatial-scenarios}. All scenarios used the same 40 positions on an otherwise empty table, giving 22 border and 18 interior plants. The air was either uniform, with the reference record applied at every position, or the prescribed field of Section~\ref{sec:greenhouse_microclimate_field}, with its longitudinal temperature gradient, the humidity derived from it and the canopy carbon dioxide depletion. Radiation was traced either for each plant alone, which kept its self-shading but excluded the other 39 plants, or with the neighbors present as occluders. The plants were either identical or variable, with persistent lognormal differences in initial size, morphology and selected physiological parameters, and the same individuals were paired between S2 and S4. Comparing S5 with S1 therefore isolates the air field, S3 with S5 the neighbors, and S2 with S1 the individual variation, all expressed as percentages of S1.

\begin{table*}[!tbp]
\centering
\caption{Scenarios of the 40-plant block and their day-30 total dry weight. Each row contains 40 simulated plants, and CV is the coefficient of variation among them.}
\label{tab:spatial-scenarios}
\small
\begin{tabular}{llllrr}
\toprule
Code & Air & Radiation tracing & Plants & Mean (g plant\(^{-1}\)) & CV (\%)\\
\midrule
S1 & Uniform & Alone & Identical & 6.481 & 0.00\\
S2 & Uniform & Alone & Variable & 6.348 & 17.41\\
S3 & Prescribed field & With neighbors & Identical & 6.801 & 5.35\\
S4 & Prescribed field & With neighbors & Variable & 6.644 & 17.66\\
S5 & Prescribed field & Alone & Identical & 8.252 & 0.26\\
\bottomrule
\end{tabular}
\end{table*}

\begin{figure*}[htbp]
    \centering
    \includegraphics[width=\textwidth]{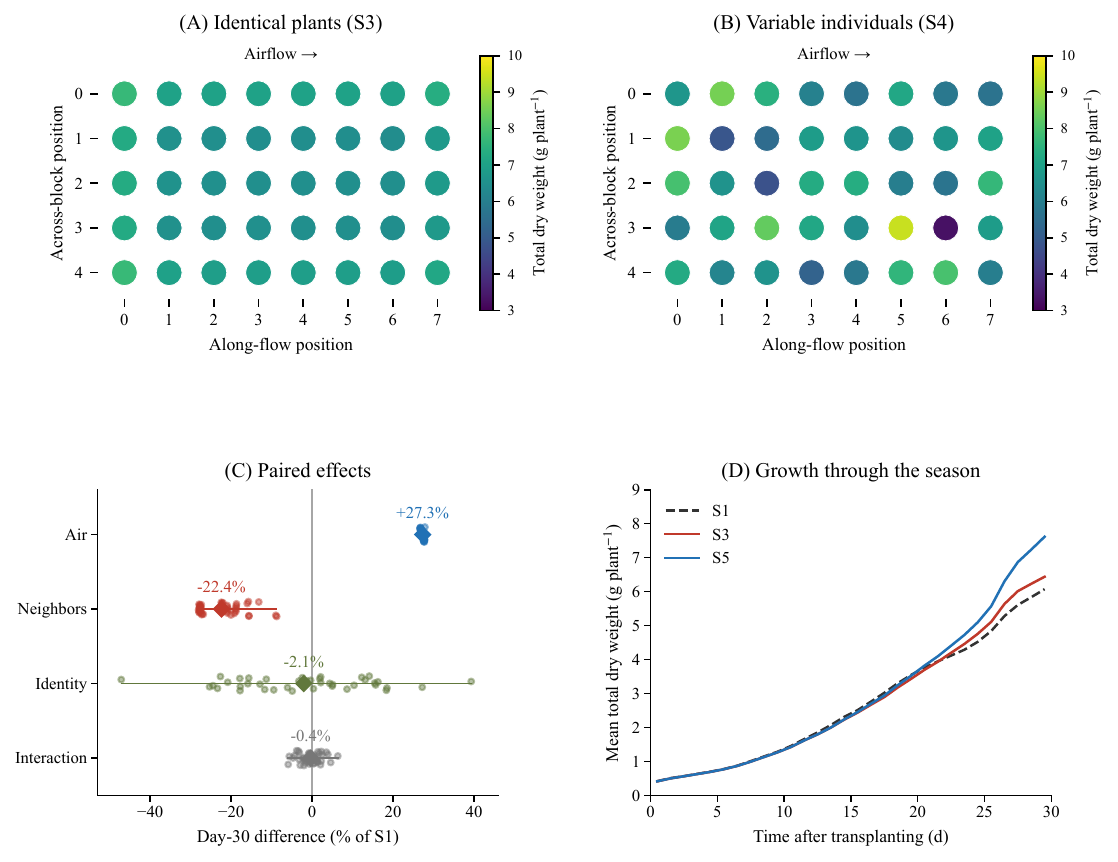}
    \caption{Growth across the simulated 40-plant block. (A, B) Day-30 dry weight of identical plants (S3) and of variable plants (S4), both with the prescribed air field and neighboring plants. (C) Paired effects of the air field, neighbor shading and plant-to-plant variation, as percentages of the uniform-air, identical-plant reference (S1). (D) Mean daily dry weight of identical plants in the reference (S1), with the field and neighbors (S3), and with the field but traced alone (S5). Scenario settings are in Table~\ref{tab:spatial-scenarios}.}
    \label{fig:result4-spatial}
\end{figure*}

Fig.~\ref{fig:result4-spatial} shows the day-30 dry weight of the block under these scenarios. In the reference scenario (S1), identical plants grew in uniform air and each was traced alone, so all of them reached the same day-30 dry weight of 6.48~g plant\(^{-1}\). When the air field and the neighboring plants were included (S3), the block mean rose to 6.80~g plant\(^{-1}\), but a clear border pattern appeared. Those 22 border plants averaged 7.08~g plant\(^{-1}\) and the 18 interior plants 6.46~g plant\(^{-1}\) (Fig.~\ref{fig:result4-spatial}A) so that interior plants accumulated 8.6\% less dry weight than border plants from identical initial states, with a coefficient of variation (CV) of 5.4\% among otherwise identical individuals. The position effect followed radiation interception. In S3, day-30 dry weight and season-mean absorbed shortwave radiation per unit leaf area had a Pearson correlation of 0.97 across the 40 positions (Fig.~\ref{fig:s-spatial}), whereas the ranges of season-mean air temperature and carbon dioxide across the block were only 0.14~\(^{\circ}\mathrm{C}\) and 0.19~ppm, so shading by neighboring leaves, not the air, produced the pattern under these prescribed conditions. Edge plants intercepting more light than interior plants is a familiar feature of lettuce stands \citep{tei1996groundcover}, and resolving it requires the arrangement of organs that determines the radiation available to each surface \citep{vos2010fspm,bailey2021resolution}. Adding plant-to-plant variation increased the spread of final weight. The CV was 17.4\% for variable plants traced alone (S2) and 17.7\% with the field and the neighbors (S4), against 5.4\% for the identical plants of S3, and in S4 the dry weight ranged from 3.31 to 9.41~g plant\(^{-1}\). In that combined scenario the correlation with absorbed radiation fell to 0.24, while initial size was strongly associated with final weight (\(r=0.89\) in S2), so the final biomass of a plant reflected both its own development and its exposure. The similar CV of S2 and S4 does not mean that the neighbors had no effect, since gains at some positions were offset by losses at others (Fig.~\ref{fig:s-spatial}).

Separating the air field from the neighbor shading revealed two opposing contributions (Fig.~\ref{fig:result4-spatial}C). With the field retained but each identical plant traced alone (S5), mean dry weight increased to 8.25~g plant\(^{-1}\), 27.3\% above S1. Adding the neighbors to that field reduced weight by an amount equivalent to 22.4\% of S1, and the shading reduced final weight at every position, with larger losses in the interior, so the +4.9\% mean difference between S3 and S1 concealed a substantial shading cost. The gain from the prescribed field came from cooler and more humid air at the crop. In the field, air temperature decreased along the house and relative humidity rose at approximately the same vapor pressure, which increased the area produced per unit dry matter through the humidity term of the specific-leaf-area relationship. Leaf area diverged from the reference before biomass did, and the block-mean S5 trajectory exceeded S1 only in the final third of the season (Fig.~\ref{fig:result4-spatial}D), consistent with greater expansion followed by greater radiation capture (\ref{app:airfield}). Because the longitudinal gradient and the humidity relation were prescribed inputs rather than measured conditions, this gain is a response to those assumptions and not an observed benefit of a position in the greenhouse, and greenhouse gradients depend on ventilation and on the crop itself \citep{kittas2003gradients}. Supplied with the same sampled climate, a four-state plant-level formulation changed its day-30 dry weight by 0.153\%, against 27.3\% here (\ref{app:airfield}). Under these prescribed conditions the plant-level formulation responded by well under 2\% while the leaf-resolved formulation responded by 27.3\%, which indicates that a formulation without the expansion-to-light-capture feedback can substantially underrepresent the cumulative effect of a spatially varying environment. Because the gradient was prescribed rather than measured, the magnitude of this difference is scenario-dependent.

\begin{figure*}[htbp]
    \centering
    \includegraphics[width=\textwidth]{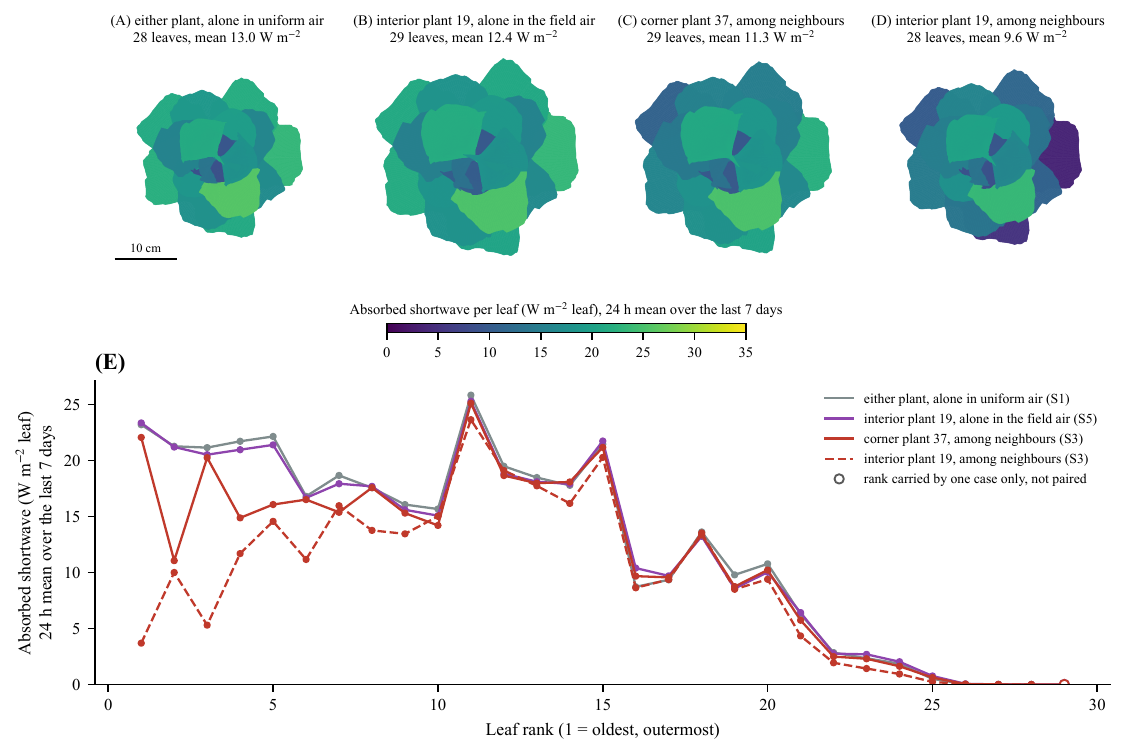}
    \caption{Absorbed shortwave radiation by leaf over the final seven days. (A) A plant in uniform air traced alone (S1). (B) The interior plant in the prescribed air field traced alone (S5). (C, D) Corner and interior plants in the same field with neighbors (S3). (E) Radiation by leaf rank for these plants. All leaf views share one color scale.}
    \label{fig:result4-leaves}
\end{figure*}

Fig.~\ref{fig:result4-leaves} shows where the spatial differences occurred within the plant. Relative to the plant traced alone in uniform air, the interior plant with neighbors received 34 to 84\% less radiation on its five oldest outer leaves, while ranks 10 to 18 changed by only +0.2 to \(-9.1\)\%. The outer leaves, which lie flattest and reach farthest into the neighboring canopy, carried almost the whole shading cost, and the enclosed inner leaves were already dark in every scenario. A block mean of a few percent therefore hides changes of tens of percent in the carbon supply of individual leaves, and the leaf is the level at which shading acts. The youngest inner leaves contributed less than 2\% of the absorbed power of the plant, but their enclosure matters for processes tied to low transpiration.

The size of the position effect also compares well with spacing experiments. Increasing the planting distance of romaine lettuce from 6.3 to 8.8~cm raised fresh mass by 46 to 62\% \citep{samy2025spacing}. At the 20~cm spacing simulated here, an 8.6\% border-to-interior difference is the smaller effect expected from wider spacing, and the model produced it from the geometry of the neighbors alone, since the 40 plants were surrounded by empty table positions and no measured gradient was imposed. The same plant therefore grew differently depending on where it stood, because the light reaching its outer leaves depended on its neighbors, and the air field and the neighbors acted in opposite directions of comparable size.

\subsection{Tipburn susceptibility and allocation sensitivity}
\label{sec:result_tipburn}

\begin{figure*}[htbp]
    \centering
    \includegraphics[width=\textwidth]{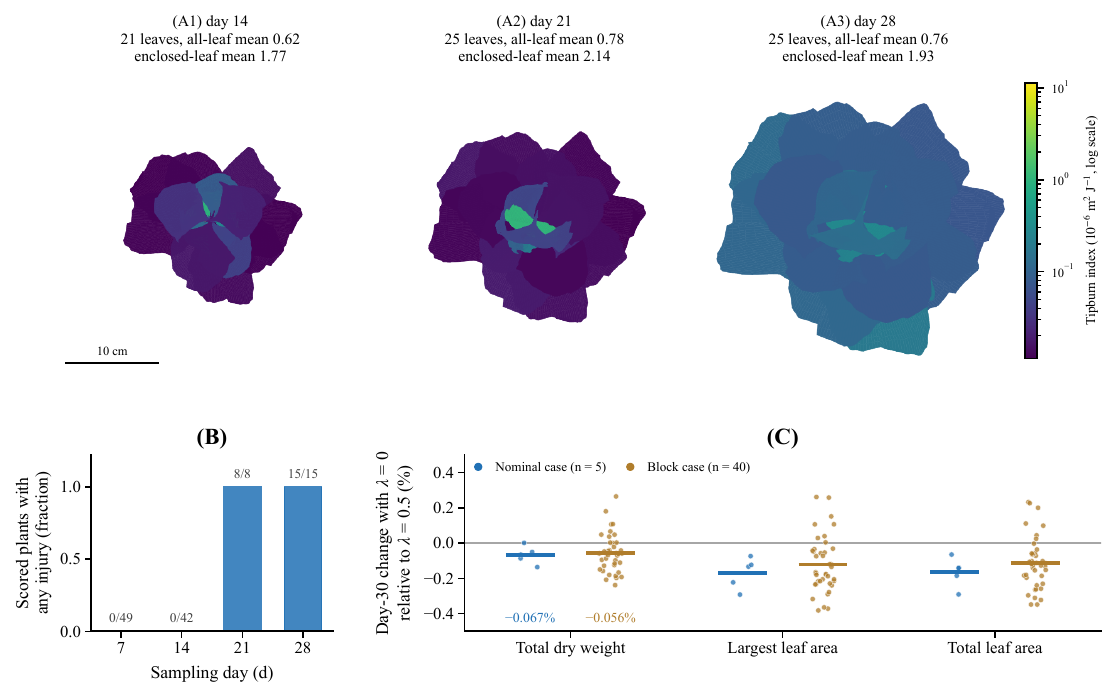}
    \caption{Tipburn susceptibility and allocation sensitivity. (A1 to A3) One plant of the variable 40-plant block on days 14, 21 and 28, with each leaf colored by the tipburn index on a logarithmic scale and the enclosed-leaf mean given for each day. (B) Fraction of scored greenhouse plants with tipburn at each sampling day. (C) Paired day-30 change in dry weight, largest-leaf area and total leaf area when the local carbon-supply rule is removed (\(\lambda=0\) against 0.5), for the five-plant case and the 40-plant block.}
    \label{fig:result5-summary}
\end{figure*}

Fig.~\ref{fig:result5-summary} summarizes two further analyses, which are reported in full in \ref{app:supporting}. The first asked whether tipburn risk emerges in the digital plant during the same period in which tipburn appeared on the real plants. Tipburn develops when calcium cannot reach the young leaves in the center of the head. Calcium moves with water, and a leaf draws water only when it transpires, so a young leaf that is enclosed by older leaves and grows fast can run short of it. The index of Section~\ref{sec:tipburn-risk-index} captures this by dividing the expansion rate of a leaf by the water it transpires \citep{collier1983calcium}. In the simulated plants the index was highest for the young enclosed leaves and lowest for the expanded outer leaves at every sampling day (Fig.~\ref{fig:result5-summary}A), with an enclosed-leaf mean of 1.93~\(\times10^{-6}\)~m\(^{2}\)~J\(^{-1}\) against 0.099~\(\times10^{-6}\)~m\(^{2}\)~J\(^{-1}\) for the outer leaves on day 28, and across the 40 plants of the variable block the median enclosed-leaf index rose from 0.47 on day 7 to 1.78, 2.14 and 3.70 on days 14, 21 and 28 as the head closed. In the greenhouse, none of the plants showed any tipburn on the early days (7th and 14th), and it appeared on plants at the later stage, the 21st and 28th day (Fig.~\ref{fig:result5-summary}B), the same timing reported in indoor trials \citep{ertle2023cultivar}. The rise of the index in the digital plant and the appearance of injury on the real plants therefore fell in the same window of head closure, between the second and the third week after transplanting. Because the index increases monotonically as the head closes, this agreement in timing is by itself weak evidence. The discriminating result is the within-plant contrast: on day 28 the enclosed leaves carried an index approximately twenty times that of the outer leaves, and the index was essentially uncorrelated with plant-level absorbed radiation across the block (Fig.~\ref{fig:s-tipburn-table}), so it tracked leaf enclosure rather than plant size or exposure. Because the measured plants were heavier than the simulated ones on the same day (Tables~\ref{tab:tipburn-correlations} and~\ref{tab:tipburn-coverage}), the comparison was made on timing and leaf position, which is what the index is meant to give.

The second analysis tested one assumption inside the allocation rule. When the plant divides its new dry matter among the leaves, the share of each leaf depends on its age and size, and on top of that the model gives a slightly larger share to a leaf that receives more light, because sugar is made in that leaf and tends to be used near where it is made. This local carbon-supply term was added from the physiology rather than fitted to data, so it was necessary to know whether the results of this paper depended on it. The same plants were therefore simulated with the term switched off, \(\lambda=0\) instead of 0.5, with everything else unchanged. Mean day-30 dry weight changed by \(-0.067\)\% in the five-plant case and \(-0.056\)\% in the 40-plant block, and no trait of any plant changed by more than 0.4\% (Fig.~\ref{fig:result5-summary}C). The results reported above therefore do not rest on this term, and it is kept because it represents the movement of carbon toward the leaves that assimilate more \citep{minchin2005phloem}.

%% file: sections/supplement_results_4_6.tex
\suppsec{Protocol of the simulation experiments}
\label{app:protocol}

Spatial growth was simulated over 30 days at 40 positions arranged as eight channels by five holes at 0.2032~m spacing along the airflow direction. The remaining table positions were empty, giving 22 border and 18 interior plants. Five scenarios separated the prescribed air field, the neighboring plants and the assigned individual variation (Table~3 of the main text). Plants traced alone retained self-shading but excluded the other plants as occluders. Paired differences used the same positions, and all percentage contrasts were normalized by the uniform-air, identical-plant reference (S1). Individual variation combined persistent differences in initial state and morphology, and means, ranges and coefficients of variation describe the simulated population.

The air field applied a longitudinal temperature gradient of 0.1~\(^{\circ}\mathrm{C}\)~m\(^{-1}\), referenced to the greenhouse center and present during both light and dark periods. Relative humidity was derived from the local temperature at a common vapor pressure, with saturation clipping. Canopy carbon dioxide depletion was 20~ppm over 0.4~m, scaled with radiation, and air speed was 0.09~m~s\(^{-1}\). These values follow the literature profiles of Section~2.3.2 of the main text. Season-mean absorbed irradiance per unit leaf area and the final-week, 24~h mean absorbed power per plant are reported separately.

The tipburn comparison asked whether the risk index rises in the digital plant during the same period in which tipburn appeared on the real plants. Leaves with a blade area of at least 1~cm\(^2\) were counted and the youngest third, rounded up, defined the enclosed group. The plant index was the mean enclosed-leaf index over the seven days before each weekly sampling. Observed severity was summarized within each sampling day, with missing scores treated as missing. Spearman correlations described the association with measured size within a day, because pooling ages would mix the joint increase of size and injury with age. The range of measured plant sizes was compared with the simulated range for each sampling day.

The local carbon-supply modifier was evaluated by paired runs with \(\lambda=0.5\) and \(\lambda=0\) in the five-plant case and in the variable 40-plant block. Initial states, positions, identities, forcing and numerical settings were held fixed and no parameter was refitted. Each run evolved its own geometry. Relative changes in day-30 traits used the \(\lambda=0.5\) value as denominator, and leaf-profile comparisons paired persistent leaf identities.

\suppsec{Spatial scenarios and carbon balance}
\label{app:spatial}

\begin{figure*}[!tbp]
\centering
\includegraphics[width=\textwidth,height=.74\textheight,keepaspectratio]{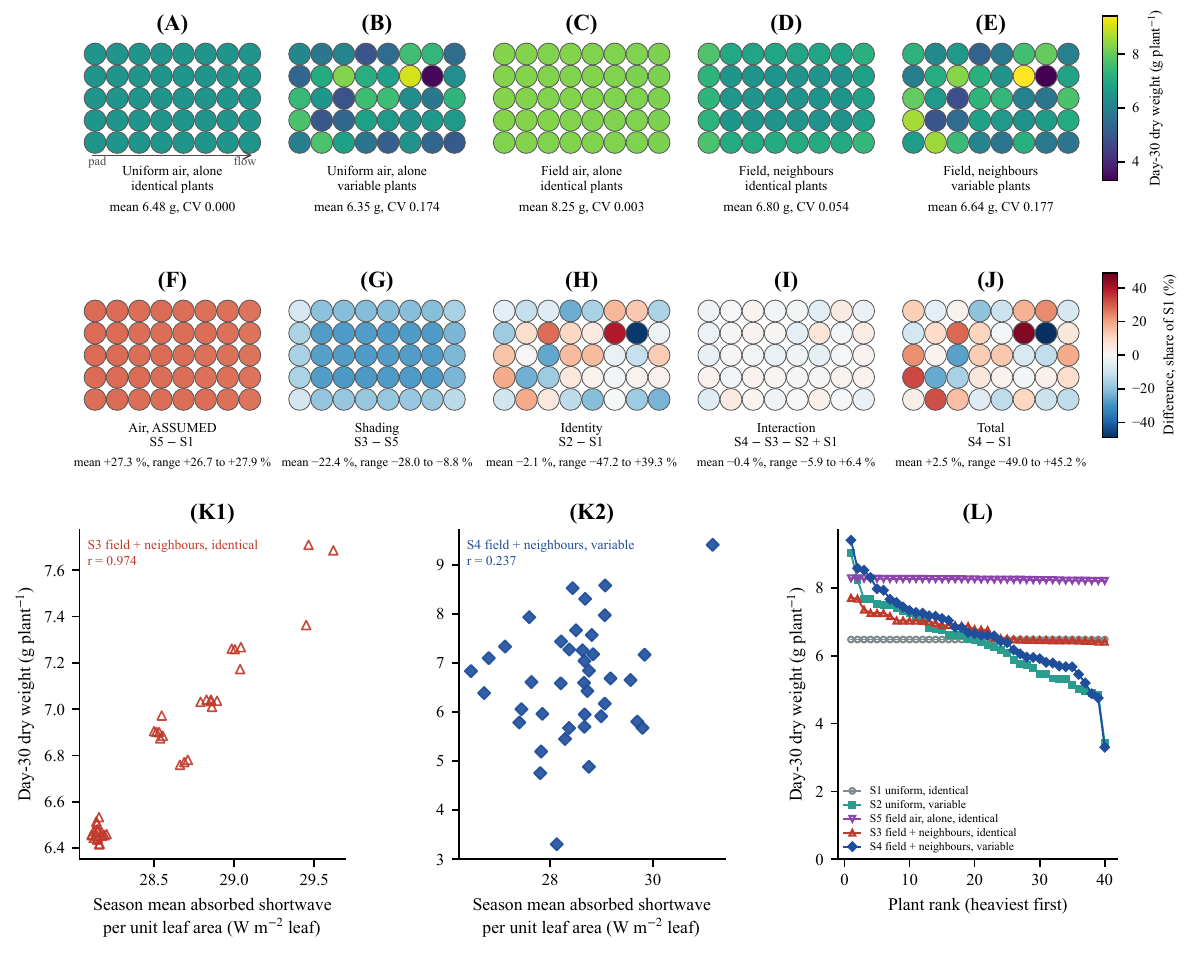}
\caption{Complete spatial scenario comparison. (A to E) Day-30 total dry weight for S1, S2, S5, S3 and S4 on a common color scale. (F to J) Per-plant air-field, neighbor, identity, interaction and total contrasts, normalized by S1 and drawn on a common diverging scale. (K1, K2) Dry weight against season-mean absorbed shortwave per unit leaf area for S3 and S4 with their Pearson correlations. (L) Sorted plant weights in each scenario. The block contains 22 border and 18 interior positions, with empty positions outside it.}
\label{fig:s-spatial}
\end{figure*}

The five scenarios used the same 40 occupied positions on an otherwise empty nutrient film technique table. Plants traced alone kept their self-shading and excluded the other 39 plants, and plants traced with neighbors included their evolving geometry. Individual variation was assigned deterministically, with the same individuals paired between S2 and S4. The air-field contrast was S5 minus S1, the neighbor contrast S3 minus S5, the identity contrast S2 minus S1, and the interaction S4 minus S3 minus S2 plus S1, all as percentages of S1 (Fig.~\ref{fig:s-spatial}). The air and neighbor contrasts sum to S3 minus S1 by construction. Expressed relative to S5, adding neighbors reduced the mean day-30 weight by 17.6\%. The identity contrast averaged \(-2.1\)\% of S1, with a range of \(-47.2\) to +39.3\%, and the interaction averaged \(-0.36\) percentage points with a mean absolute magnitude of 2.26 points.

\begin{table*}[!tbp]
\centering\small
\caption{Mean leaf area, radiation and carbon balance per plant. Area is total physiological leaf area on day 30, irradiance is season-mean absorbed shortwave per unit leaf area, and power is the final-seven-day, 24~h mean absorbed shortwave per plant. Gross assimilation and maintenance respiration are 30-day carbon integrals. Irradiance and power use different averaging windows.}
\label{tab:spatial-budget}
\begin{tabular}{lrrrrr}
\toprule
 & Area & Irradiance & Power & Gross assimilation & Respiration\\
 & (cm\(^2\)) & (W m\(^{-2}\) leaf) & (W) & (g C) & (g C)\\
\midrule
S1 & 2180.5 & 33.37 & 2.646 & 4.024 & 0.789\\
S2 & 2104.8 & 33.34 & 2.572 & 3.913 & 0.783\\
S3 & 2774.8 & 28.54 & 2.642 & 4.115 & 0.747\\
S4 & 2681.4 & 28.46 & 2.562 & 3.996 & 0.737\\
S5 & 3490.3 & 30.47 & 3.874 & 4.910 & 0.794\\
\bottomrule
\end{tabular}
\end{table*}

The air-field gain developed late in the season. In the daily block means, S5 stayed below S1 for 20 of the 30 days, with the largest deficit of \(-3.49\)\% at day 13.5 and the first surplus at day 20.5, and S3 crossed S1 at day 22.5. The plants in the field first built more leaf area and only later turned that area into more dry matter, which Table~\ref{tab:spatial-budget} shows as the largest leaf area and absorbed power in S5.

\suppsec{Diagnostic evaluation of the prescribed air-field response}
\label{app:airfield}

The temperature field was prescribed with a constant longitudinal gradient during both day and night, and relative humidity was derived from temperature at a common vapor pressure. For diagnostic plant 1, the field lowered the mean local air temperature by 1.09~\(^{\circ}\mathrm{C}\), by 1.00~\(^{\circ}\mathrm{C}\) in daylight and 1.21~\(^{\circ}\mathrm{C}\) in darkness, and raised the expansion-weighted relative humidity from 0.675 to 0.755. Under the specific-leaf-area humidity factor \(f_{RH}=[1+\beta_{RH}(RH_{\mathrm{ref}}-RH)]^{-1}\), with \(\beta_{RH}=0.912\) and \(RH_{\mathrm{ref}}=0.75\), the expansion-weighted humidity factor increased by 7.26\% and the radiation factor by 1.34\%. The field and uniform-air vapor pressures differed by 0.00009\% at the median time step and by 0.76\% in mean absolute terms, with saturation clipping in 8.6\% of the time steps, so the humidity change came almost entirely from the cooler air. This one change acted on both the leaf energy balance and the area produced per unit dry matter.

In the same plant, the leaf-area difference between S5 and S1 was +0.86\% on day 1, +13.07\% on day 11 and +61.13\% on day 30, while dry weight stayed below S1 for the first 19 days. Expansion therefore came first and the biomass advantage followed once the larger canopy captured more light.

A separate four-state diagnostic supplied the same sampled local climate to the plant-level model with its other settings fixed. Its day-30 dry weight was 10.709~g under uniform forcing and 10.726~g with the sampled field (+0.153\%). Changing temperature and carbon dioxide while keeping the original humidity gave \(-1.525\)\%, changing humidity alone gave +1.583\%, and the temperature-only and carbon-dioxide-only contrasts were +1.812\% and \(-3.618\)\%, all relative to that model's own baseline. The plant-level model therefore responded to the same air by well under 2\%, whereas the leaf-resolved model responded by 27.3\%, which indicates that the air-field gain arose from the leaf-level feedback between expansion and light capture that the plant-level model does not contain.

\suppsec{Measured tipburn and spatial index comparisons}
\label{app:tipburn}

The leaf views of the simulated index are Fig.~10 of the main text. Across the 40 plants of the variable block, the median enclosed-leaf index rose from 0.47 on day 7 to 1.78, 2.14 and 3.70~\(\times10^{-6}\)~m\(^{2}\)~J\(^{-1}\) on days 14, 21 and 28, and the identical-plant block rose from 0.41 to 3.35 over the same days. Tipburn on the real plants appeared between the day-14 and the day-21 samplings, so the rise of the simulated index and the appearance of injury fell in the same period of head closure. Only plants with recorded severity were included: 49 on day 7, 42 on day 14, eight on day 21 and 15 on day 28. All scored day-7 and day-14 plants had severity zero, and all scored day-21 and day-28 plants had positive severity. Correlations were computed separately by day, because size and injury both increase with age.

\begin{table}[htbp]
\centering\small
\caption{Within-day Spearman correlations of observed severity with measured size. Counts vary with measurement availability. The leaf-area column of the trial is the largest blade.}
\label{tab:tipburn-correlations}
\begin{tabular}{lrrrr}
\toprule
 & \multicolumn{2}{c}{Day 21} & \multicolumn{2}{c}{Day 28}\\
Measure & \(n\) & \(\rho\) & \(n\) & \(\rho\)\\
\midrule
Shoot dry weight & 8 & +0.41 & 15 & +0.37\\
Largest-leaf area & 8 & \(-0.22\) & 15 & +0.28\\
Plant diameter & 6 & \(-0.13\) & 15 & \(-0.04\)\\
\bottomrule
\end{tabular}
\end{table}

Within a sampling day, heavier plants tended to carry more severe injury, with Spearman correlations of +0.41 on day 21 and +0.37 on day 28 for shoot dry weight (Table~\ref{tab:tipburn-correlations}). One day-21 record combined 18.2 in the shoot-weight column with a largest-leaf area of 55.66~cm\(^2\), against a day-21 median of 174.6~cm\(^2\), and omitting it moves the day-21 area correlation from \(-0.22\) to +0.41. The record was retained. Leaf number was available for only one of the 23 late-stage scored plants and was not used.

\begin{table*}[!tbp]
\centering\small
\caption{Scored plants outside the simulated shoot-weight range at each sampling day, with the recorded and simulated ranges. Shoot dry weight is as recorded in the trial workbook.}
\label{tab:tipburn-coverage}
\begin{tabular}{rrrrr}
\toprule
Day & Scored plants & Outside range & Recorded range & Simulated range\\
 & \(n\) & \(n\) & (g plant\(^{-1}\)) & (g plant\(^{-1}\))\\
\midrule
7 & 49 & 18 & 0.30 to 4.00 & 0.45 to 1.28\\
14 & 42 & 26 & 0.70 to 13.00 & 1.02 to 3.01\\
21 & 8 & 8 & 7.50 to 19.10 & 1.84 to 5.14\\
28 & 15 & 14 & 5.00 to 27.00 & 2.87 to 8.17\\
\bottomrule
\end{tabular}
\end{table*}

\begin{figure*}[!tbp]
\centering
\includegraphics[width=\textwidth,height=.72\textheight,keepaspectratio]{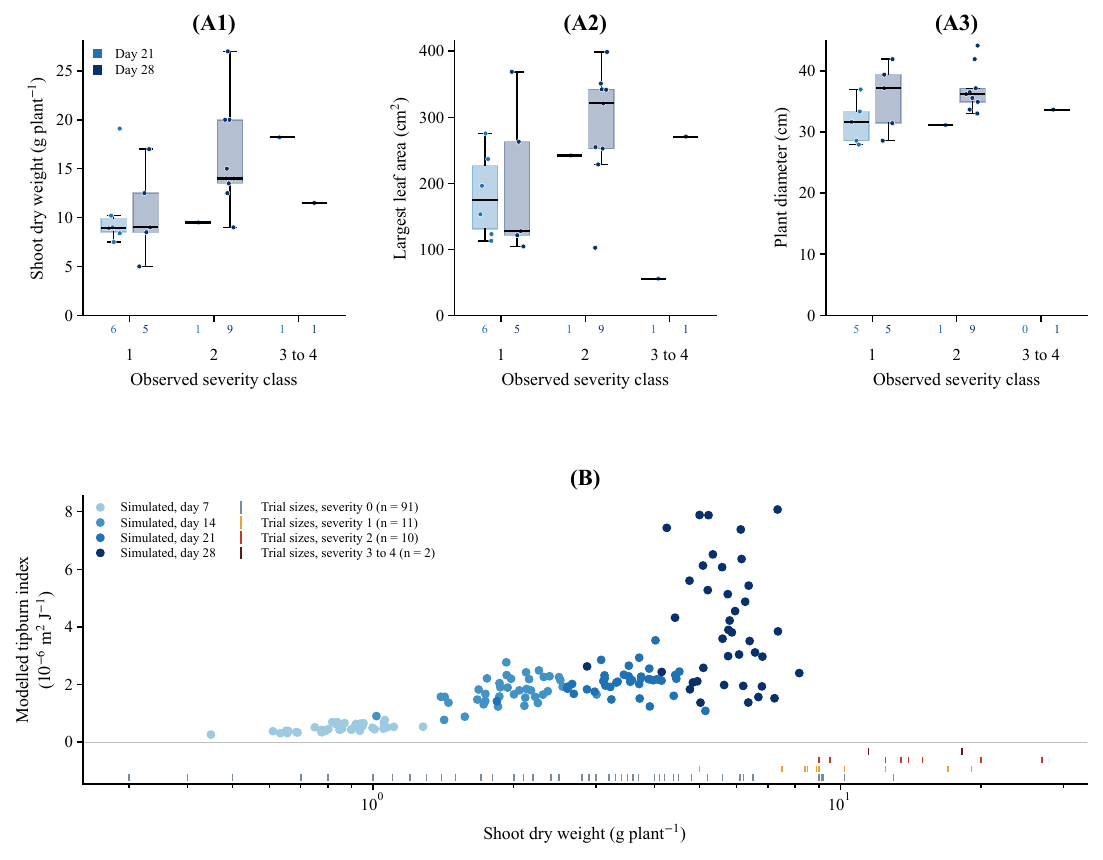}
\caption{Observed severity and plant size, with the simulated index. (A1 to A3) Measured shoot dry weight, largest-leaf area and diameter by severity class on days 21 and 28. Boxes show medians and quartiles, points show individual records, and numbers below the groups give sample counts. Classes 3 and 4 are grouped for display. (B) Simulated enclosed-leaf index against shoot dry weight for the 40 S4 plants at each sampling day, with the measured plants shown as rug marks at their recorded sizes, colored by observed severity.}
\label{fig:s-observed-tipburn}
\end{figure*}

The measured plants were heavier than the simulated ones on the same day, with 66 of the 114 scored plants outside the simulated shoot-weight range (Table~\ref{tab:tipburn-coverage}). Fig.~\ref{fig:s-observed-tipburn} therefore shows the simulated index against simulated shoot weight and the measured plants at their recorded sizes, so that the comparison is made on timing and on which leaves carry the risk. The simulated enclosed-leaf index was averaged over the seven days before each sampling date and is expressed in \(10^{-6}\)~m\(^2\)~J\(^{-1}\).

At day 28, the mean index was 3.88 in the identical-plant block and 4.04 in the variable block, with CVs of 44.4\% and 50.6\%, and the plants ranged from 1.99 to 8.66 and from 1.37 to 8.08 in the same units. Identical plants had much smaller CVs on days 14 and 21 (3.2\% and 9.0\%), so the spread of the index across the block appeared as the canopies developed. The Spearman correlation of the index with season-mean absorbed light was +0.57, \(-0.15\) and +0.27 on days 14, 21 and 28 in S3 and \(-0.09\), \(-0.07\) and \(-0.07\) in S4, so the index followed the enclosure of the young leaves rather than the light on the plant.

\begin{figure*}[!tbp]
\centering
\includegraphics[width=\textwidth,height=.55\textheight,keepaspectratio]{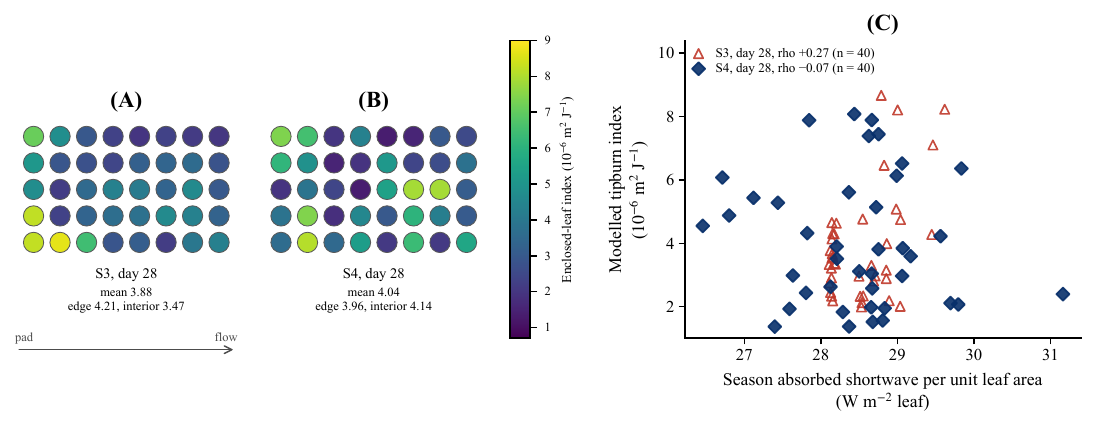}
\caption{Day-28 simulated enclosed-leaf index across the 40-plant block. (A) Identical plants (S3). (B) Variable individuals (S4). (C) The same index against season-mean absorbed shortwave per unit leaf area, with within-scenario Spearman correlations. Both maps use one linear color scale, shared with Fig.~\ref{fig:s-tipburn-early}, in \(10^{-6}\)~m\(^2\)~J\(^{-1}\). Map annotations give whole-block, border and interior means. Air speed was 0.09~m~s\(^{-1}\).}
\label{fig:s-tipburn-table}
\end{figure*}

\begin{figure*}[!tbp]
\centering
\includegraphics[width=\textwidth,height=.55\textheight,keepaspectratio]{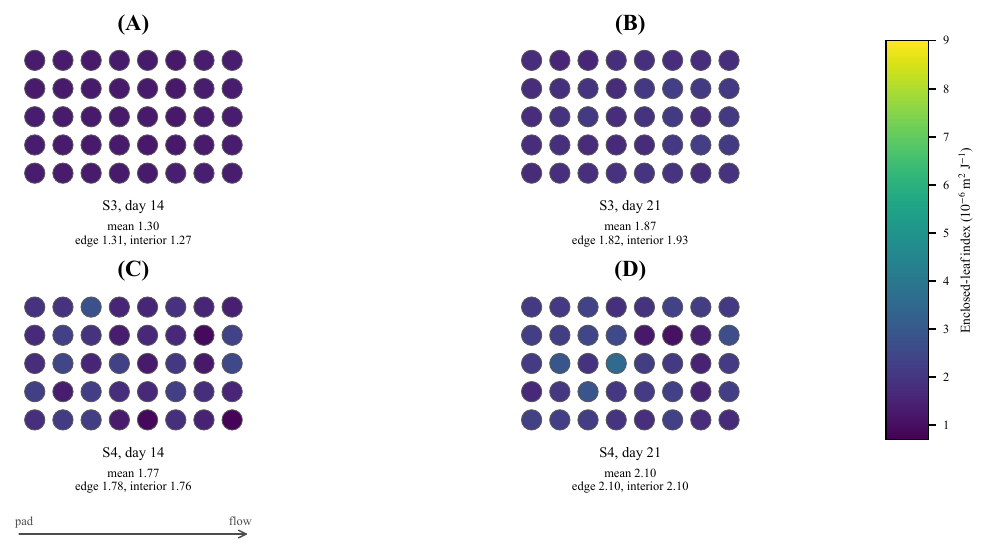}
\caption{Simulated enclosed-leaf index maps for the same block on days 14 and 21, for identical plants (A and B, S3) and variable individuals (C and D, S4). The linear color scale is the same as in Fig.~\ref{fig:s-tipburn-table}. The near-uniform early S3 maps show the smaller spread at these dates, and the printed means allow comparison without rescaling the colors.}
\label{fig:s-tipburn-early}
\end{figure*}

\suppsec{Paired allocation ablation and leaf-profile sensitivity}
\label{app:ablation}

The nominal and block comparisons contained five and 40 paired plants. The coefficient \(\lambda\) controlling the local carbon-supply contribution was changed from 0.5 to zero, keeping the common growth pool, the developmental allocation relationships and the expansion-headroom factor \(\kappa=1\). All 45 paired initial states and positions were identical, and \(\lambda\) was the only changed setting. Geometry evolved independently in each run, so the final differences include its feedback on radiation capture.

\begin{table*}[!tbp]
\centering\small
\caption{Day-30 ablation results. Changes are \(\lambda=0\) minus \(\lambda=0.5\), and relative changes are averaged over the paired plant percentages rather than calculated from rounded group means.}
\label{tab:ablation-traits}
\begin{tabular}{lrr}
\toprule
Measure & Nominal (5 plants) & Block (40 plants)\\
\midrule
Mean total dry weight, \(\lambda=0.5\) (g) & 7.81318 & 6.64383\\
Mean total dry weight, \(\lambda=0\) (g) & 7.80766 & 6.63974\\
Mean paired dry-weight change (\%) & \(-0.0671\) & \(-0.0560\)\\
Range of dry-weight change (\%) & \(-0.1358\) to +0.0018 & \(-0.2380\) to +0.2653\\
Mean paired largest-leaf-area change (\%) & \(-0.1687\) & \(-0.1218\)\\
Mean paired total-leaf-area change (\%) & \(-0.1641\) & \(-0.1128\)\\
Mean initiated-leaf-number change (leaves) & \(-0.0049\) & \(-0.0106\)\\
\bottomrule
\end{tabular}
\end{table*}

For a plant with reference leaf areas \(A_i\) and ablated areas \(A'_i\), the profile statistic was \(100\sqrt{n^{-1}\sum_i[(A'_i-A_i)/A_i]^2}\), evaluated over reference leaves meeting the stated area threshold and then averaged across plants. It gives each included leaf equal weight, so percentage changes in small leaves dominate it. The largest block profile value (11.82\%) came from a leaf declining from 2.936 to 1.239~cm\(^2\), a 57.8\% change on 1.696~cm\(^2\) of area. Mean absolute dry-weight differences were similar between interior and border plants (0.098\% and 0.100\%), while the mean profile difference was higher in the interior (2.24\% against 1.24\%), where more small leaves sit in shade.

\begin{table*}[!tbp]
\centering\small
\caption{Leaf-profile and absolute-area sensitivity to removing the modifier. Profile RMS values are mean plant-level root mean square percentage differences. Absolute-area quantities use paired reference leaves of at least 1~cm\(^2\), with gains and losses retained.}
\label{tab:ablation-profiles}
\begin{tabular}{lrr}
\toprule
Measure & Nominal & Block\\
\midrule
Profile RMS, all positive reference areas (\%) & 1.209 & 1.098\\
Profile RMS, reference area \(\geq1\) cm\(^2\) (\%) & 1.814 & 1.690\\
Profile RMS, reference area \(\geq10\) cm\(^2\) (\%) & 0.656 & 0.714\\
Mean absolute area-change sum (cm\(^2\) plant\(^{-1}\)) & 8.879 & 6.765\\
Mean signed area change (cm\(^2\) plant\(^{-1}\)) & \(-5.496\) & \(-3.138\)\\
Mean offsetting gains/losses (cm\(^2\) plant\(^{-1}\)) & 1.692 & 1.814\\
Offsetting component / mean reference area (\%) & 0.051 & 0.068\\
\bottomrule
\end{tabular}
\end{table*}

The offsetting component was calculated per plant as \([\sum_i|\Delta A_i|-|\sum_i\Delta A_i|]/2\), where \(\Delta A_i=A'_i-A_i\). It measures area gained by some leaves and lost by others, and at 0.05 to 0.07\% of the reference area it shows that the modifier moved very little area between leaves. Dry weight, largest-leaf area and total leaf area were therefore insensitive to the modifier in both cases, and the profile differences came from small shaded leaves rather than from a redistribution of the canopy.

\begin{figure*}[!tbp]
\centering
\includegraphics[width=\textwidth,height=.73\textheight,keepaspectratio]{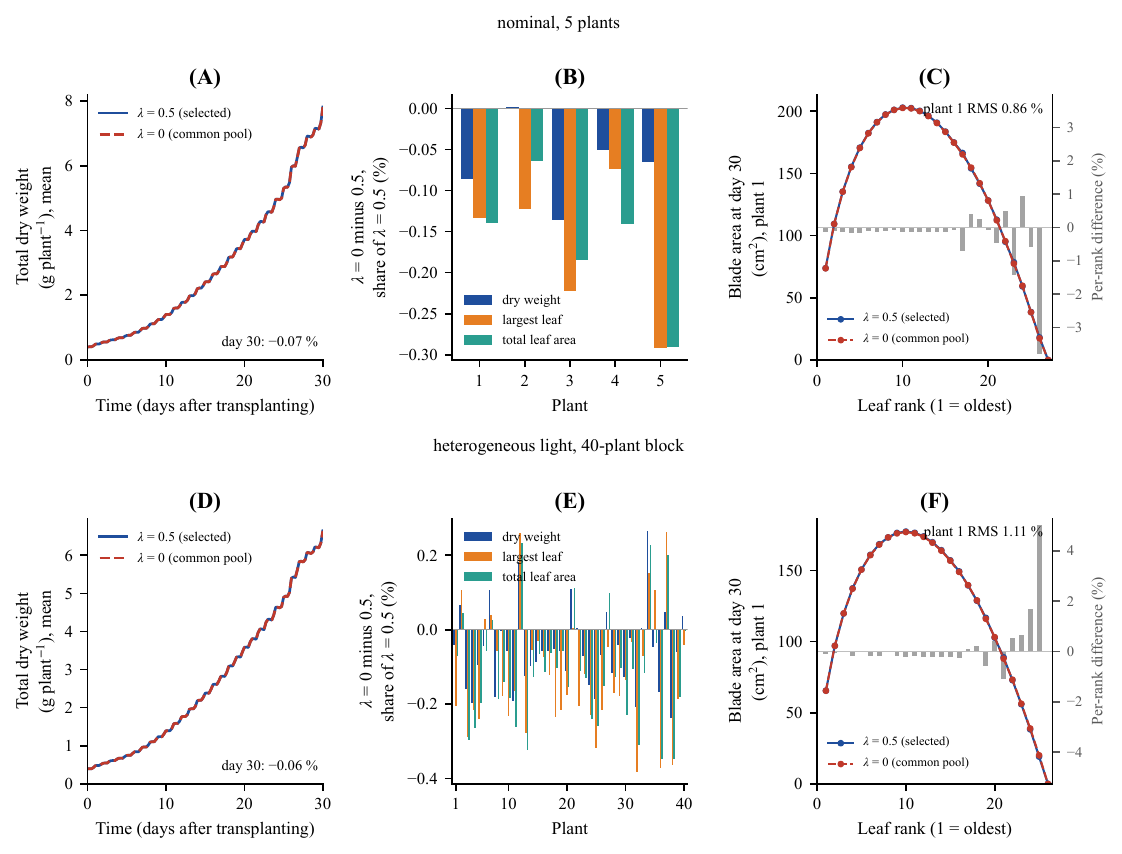}
\caption{Local carbon-supply ablation for the nominal five-plant case (top) and the variable 40-plant block (bottom). Left: mean total dry weight with \(\lambda=0.5\) and \(\lambda=0\). Middle: paired day-30 differences in dry weight, largest-leaf area and total leaf area, relative to the \(\lambda=0.5\) value. Right: individual-leaf area profiles for plant 1 in each case, with relative differences shown as bars on the secondary axis for reference areas of at least 1~cm\(^2\). The annotated profile RMS values describe the illustrated plant rather than the case averages in Table~\ref{tab:ablation-profiles}. No parameter was refitted.}
\label{fig:s-ablation}
\end{figure*}